\documentclass[conference]{ieeeconf}  

\IEEEoverridecommandlockouts                              

\usepackage{cite}
\usepackage{mathtools}
\usepackage{amsmath,amssymb,amsfonts}
\usepackage{algorithmic}
\usepackage{graphicx}
\usepackage{subfigure}
\usepackage{verbatim}
\usepackage{textcomp}
\usepackage{float}
\usepackage{esvect}
\usepackage{multirow}
\usepackage{tabularx}
\usepackage[utf8]{inputenc}
\usepackage{diagbox}
\usepackage{color}

\usepackage{geometry}

\title{\LARGE \bf
EPOSIT: An Absolute Pose Estimation Method for Pinhole and Fish-Eye Cameras
}

\author{Zhaobing Kang, Wei Zou, Zheng Zhu, Chi Zhang and Hongxuan Ma
\thanks{Zhaobing Kang, Zheng Zhu, Chi Zhang and Hongxuan Ma are affiliated to Institute of Automation Chinese Academy of Sciences, University of Chinese Academy of Sciences;
        { \tt\small kangzhaobing2017@ia.ac.cn}}%
\thanks{Wei Zou is affiliated to Institute of Automation, Chinese Academy of Sciences, Beijing China and TianJin Intelligent Tech.Institute of CASIA Co. ,Ltd, Tianjin China;
        {\tt\small wei.zou@ia.ac.cn}}
\thanks{This work is supported by The National Key Research and Development Program of China (Project 2017YFB1300104), and The National Natural Science Foundation of China (Grant No. 61773374).}}%
\begin{document}

\maketitle
\thispagestyle{empty}
\pagestyle{empty}

\begin{abstract}

This paper presents a generic 6DOF camera pose estimation method, which can be used for both the pinhole camera and the fish-eye camera. Different from existing methods, relative positions of 3D points rather than absolute coordinates in the world coordinate system are employed in our method, and it has a unique solution. The application scope of POSIT (Pose from Orthography and Scaling with Iteration) algorithm is generalized to fish-eye cameras by combining with the radially symmetric projection model. The image point relationship between the pinhole camera and the fish-eye camera is derived based on their projection model. The general pose expression which fits for different cameras can be acquired by four noncoplanar object points and their corresponding image points. Accurate estimation results are calculated iteratively. Experimental results on synthetic and real data show that the pose estimation results of our method are more stable and accurate than state-of-the-art methods. The source code is available at https://github.com/k032131/EPOSIT.

\end{abstract}

\section{INTRODUCTION}
Estimating the camera pose using minimal 3D-2D point correspondences is an important problem in geometry computer vision, i.e. the Perspective-n-Point (PnP) problem. For most conventional methods \cite{29, 30, 31, 32}, accurate camera intrinsic parameters and 3D point coordinates in the world coordinate system are required. In recent years, the problems with unknown focal length (PnPf) or with unknown focal length and radial distortion (PnPfr) are widely studied \cite{16, 17, 18, 19, 20}, but scale ambiguity and multi-solution in most solvers limit their applications.

For the PnP problem with precise camera intrinsic parameters, the pose can be estimated by at least three 3D-2D point correspondences \cite{4,5,26}. The famous POSIT algorithm is proposed in \cite{15}, where the unique solution can be iteratively calculated by four 3D-2D point correspondences. In order to make algorithms more robust to the measurement error and the image noise, multi-sensor fusion \cite{8} and optimization \cite{9, 10, 11, 12, 13, 14} methods are introduced. The major solvers for this problem are presented in \cite{6} and \cite{7}, and their robustness to image noise and numerical stability are discussed in \cite{28}. However, the above methods usually require accurate camera calibration. 

To obtain the camera pose when accurate intrinsic parameters cannot be acquired, the problems with unknown focal length (PnPf) \cite{16, 17, 18} or with unknown focal length and radial distortion (PnPfr) \cite{19, 20, 21, 22, 23, add1, add2} have been extensively studied. In \cite{18}, the focal length is estimated by a novel sampling scheme, which guides the sampling process towards promising focal length values and avoids considering all possibilities. A P3.5P method is proposed in \cite{24}, where either $x$ or $y$ coordinate of an image point is combined with P3P to estimate the camera pose with unknown focal length. In \cite{19}, the camera pose is estimated by four point correspondences, and the result is calculated by the Gr$\ddot{o}$bner basis solver. 
By adding one point, five point correspondences can reduce the number of solutions and accelerate the solving speed \cite{20}. 
In \cite{16}, a general pose solution which combines Hidden variable resultant and Gr$\ddot{o}$bner basis techniques is proposed. Different from solving an algebraic problem in \cite{16}, bivariate polynomial methods are employed in \cite{17}, and the RANSAC is introduced to make the result robust to outliers. In \cite{22}, the solving speed is dramatically improved by separating the problem into noncoplanar and planar scenes. For the PnPf problem, accurate image distortion coefficient is essential. Although PnPfr methods can estimate the camera pose with unknown radial distortion, additional constraints, such as the minimal reprojection error are needed to determine the real value from multi-solution.

Inspired by the POSIT algorithm \cite{15}, this paper presents a novel absolute pose estimation method called extended POSIT (EPOSIT) algorithm, which fits for both the pinhole camera and the fish-eye camera. In POSIT, the perspective projection model is employed to derive the camera pose expression, so it only can be used for the pinhole camera. In order to extend the POSIT algorithm to the fish-eye camera, we take the radially symmetric projection model \cite{25} into consideration, and the generic pose expression is derived based on the image point relationship between the pinhole camera and the fish-eye camera. The pose estimation results can be calculated by the iterative method. 

EPOSIT has three advantages compared with existing methods. Firstly, it can accurately estimate the pose of the pinhole camera and the fish-eye camera without precise calibration. Secondly, since the pose is estimated by the iterative method, the solution is unique. Lastly, relative positions of 3D points are employed, which is meaningful when the world coordinate system is fixed, such as visual servoing control of the mobile robot. In the mobile robot visual control, it is more convenient to directly estimate the robot pose in the camera coordinate system using points with known relative positions than absolute coordinates in the world coordinate system.  

\section{EXTENDED POSIT ALGORITHM}
\subsection{POSIT Algorithm}
The POSIT algorithm is a classic pose estimation method, where the camera pose can be estimated by four noncoplanar object points according to the perspective projection model and the scaled orthographic projection (SOP) model. These two models are shown in Fig. \ref{fig1:a}.
In this figure, $M_{0}$ and $M_{i}\ (i=1,...,3)$ represent four noncoplanar object points, whose relative positions are known, and an object coordinate system $M_{o}uvw$ is built. The camera optical center is denoted as $O$, the camera coordinate system is $Oxyz$, where the unit vectors of its three axes are denoted as $\vec{i}$, $\vec{j}$ and $\vec{k}$, and the image plane is denoted as $G$. To facilitate describing the SOP model, we draw a plane $K$ that passes through point $M_{0}$, parallels to the plane $G$ and intersects with $z$-axis at point $H$. The distance from $O$ to the plane $K$ is $Z_{0}$.

Based on the perspective projection model, $M_{0}$ and $M_{i}$ are respectively projected to image points $m_0$ and $m_{i}$. The line of sight for $M_{i}$ intersects with the plane $K$ at point $N_{i}$.
The SOP model is an approximation to the perspective projection model. In SOP, $M_{0}$ and $M_{i}$ are assumed at the same depth $Z_{0}$. Therefore, $M_{i}$ is orthographically projected onto the plane $K$ at $P_{i}$, whose corresponding image point is ${p}_{i}$.
For the POSIT algorithm, the problem to be solved can be described as below.

\textbf{Problem}: Given the camera focal length $f$, four noncoplanar object points $M_0$...$M_3$, their relative positions and corresponding image points $m_0$...$m_3$, estimating the pose of the frame $M_0uvw$ relative to the camera frame $Oxyz$.

{\setlength{\abovecaptionskip}{-2pt}
\setlength{\belowcaptionskip}{-2pt}
\begin{figure} \centering 
\subfigure[] { \label{fig1:a} 
\includegraphics[scale=0.29]{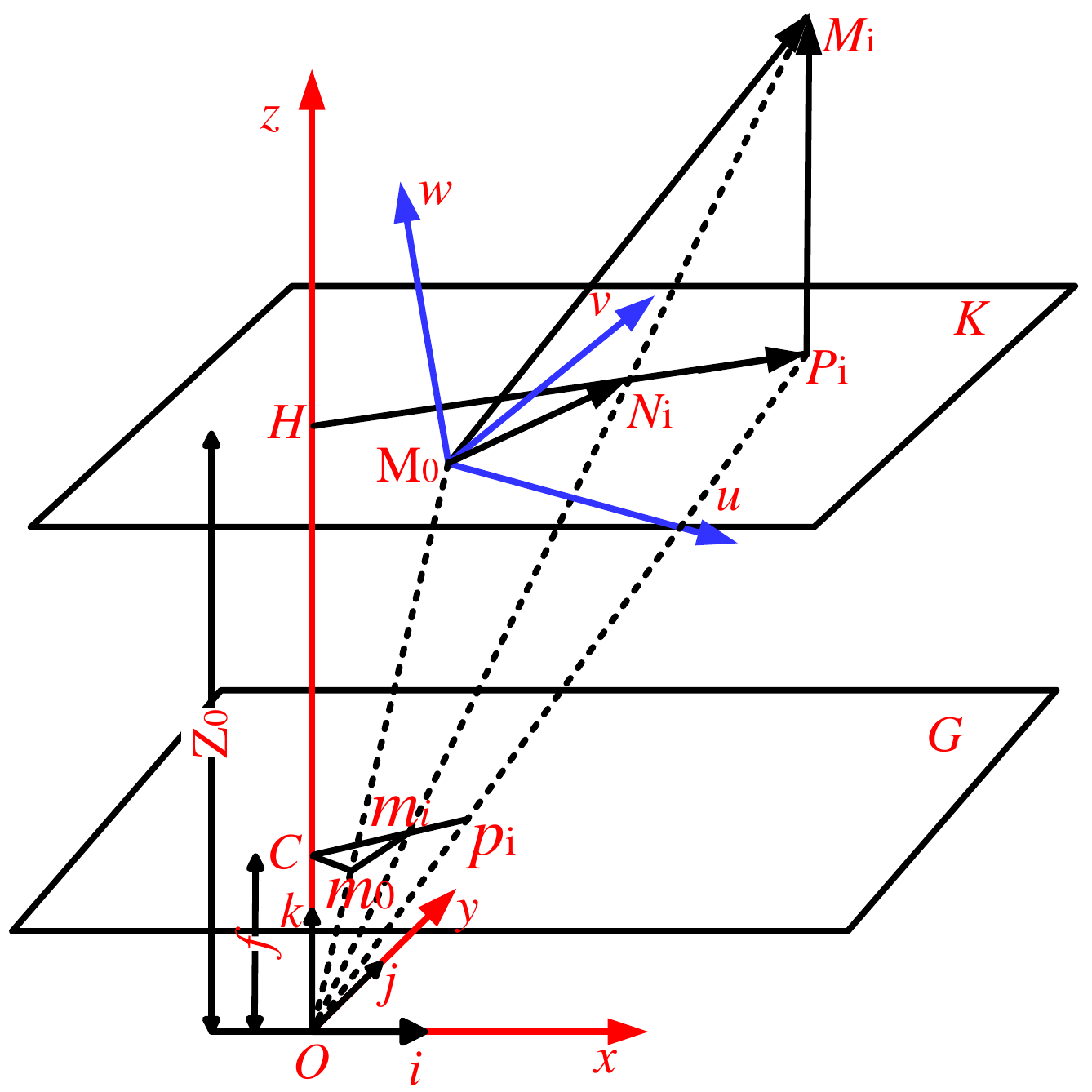} 
} 
\subfigure[] { \label{fig1:b} 
\includegraphics[scale=0.35]{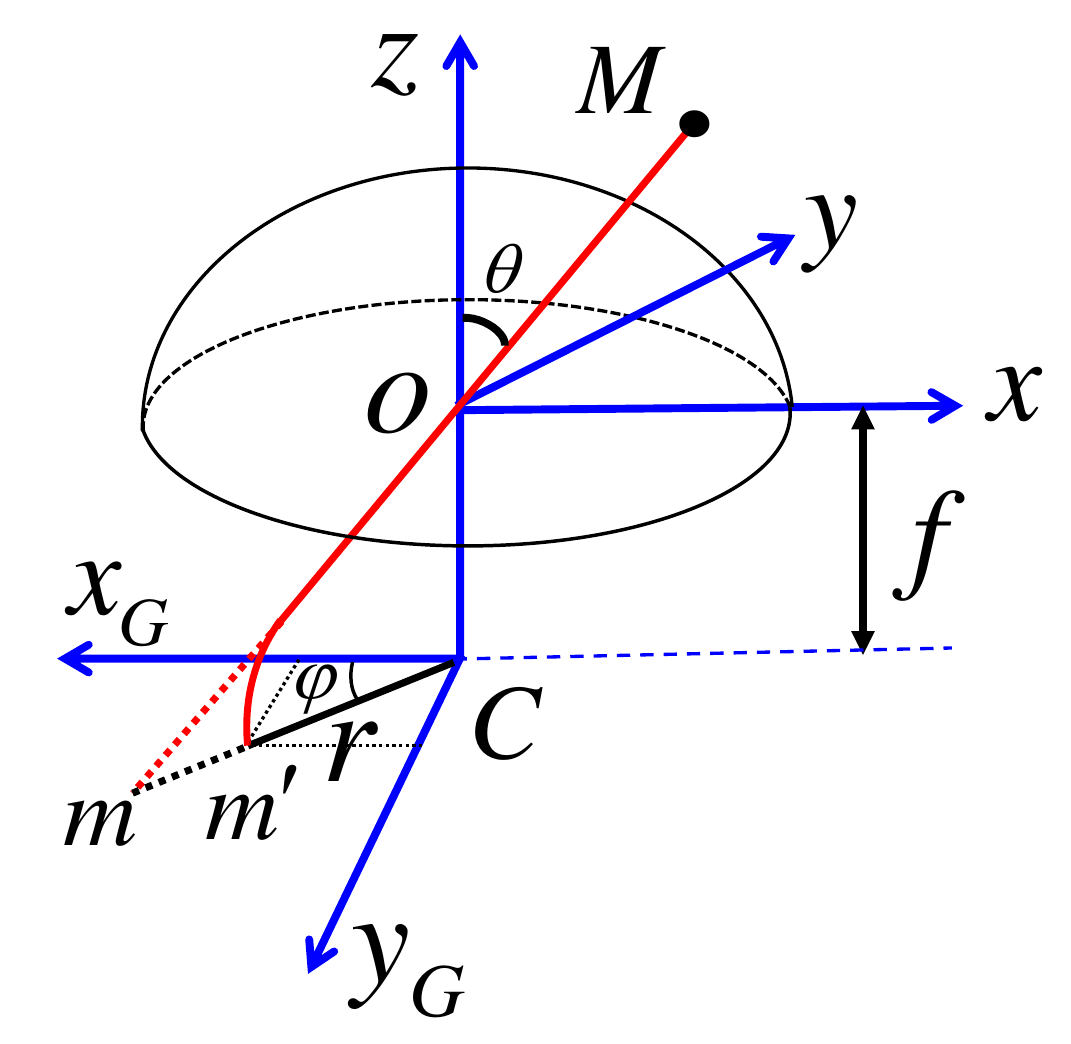}
} 
\caption{(a) The perspective projection model used for the POSIT algorithm. The object and the camera coordinate systems are denoted as $M_0uvw$ and $Oxyz$, respectively. Object points $M_0$, $M_i$ and $P_i$ are projected to image points $m_0$, $m_i$ and $p_i$. (b) The radially symmetric projection model. The object point $M$ is respectively projected to image points $m$ and $m^{'}$ for the pinhole camera and the fish-eye camera.}
\label{fig1} 
\end{figure}}

Let $\prescript{c}{}{\emph{\textbf{R}}}_{o}^{}$ and $\prescript{c}{}{\emph{\textbf{T}}}_{o}^{}$ respectively represent the rotation matrix and the translation vector of the frame $M_0uvw$ relative to the frame $Oxyz$, which  are expressed as
\vspace{-0.5em}
\begin{equation*}
\prescript{c}{}{\emph{\textbf{R}}}_{o}^{}=\begin{bmatrix}\begin{smallmatrix}i_{u} & i_{v} & i_{w}\\j_{u} & j_{v} & j_{w}\\k_{u} & k_{v} & k_{w}\end{smallmatrix}\end{bmatrix},\ \prescript{c}{}{\emph{\textbf{T}}}_{o}^{}=\protect\overrightarrow{OM_{0}}
\end{equation*}where \begin{math}i_{u}\end{math}, \begin{math}i_{v}\end{math} and \begin{math}i_{w}\end{math} are the coordinates of \begin{math}\vec{i}\end{math} in the frame $M_0uvw$.
The coordinates of points \begin{math}M_{0}\end{math}, \begin{math}M_{i}\end{math} and \begin{math}P_{i}\end{math} in the camera coordinate system are denoted as \begin{math}\left(X_{M0}^{}, Y_{M0}^{}, Z_{0}\right)^T\end{math}, \begin{math}\left(X_{Mi}^{}, Y_{Mi}^{}, Z_{i}\right)^T\end{math} and \begin{math}\left(X_{Mi}^{}, Y_{Mi}^{}, Z_{0}\right)^T\end{math} respectively. Their corresponding image point coordinates are respectively denoted as \begin{math}\left(x_{M0}^{}, y_{M0}^{}\right)^T\end{math}, \begin{math}\left(x_{Mi}^{}, y_{Mi}^{}\right)^T\end{math} and \begin{math}\left(x_{Pi}^{}, y_{Pi}^{}\right)^T\end{math}. According to the perspective projection model, these image point coordinates can be calculated by
\begin{align}
&x_{M0}^{}=X_{M0}^{}f/Z_{0},\;y_{M0}^{}=Y_{M0}^{}f/Z_{0}\label{eq7}\\
&x_{Mi}^{}=X_{Mi}^{}f/Z_{i},\;y_{Mi}^{}=Y_{Mi}^{}f/Z_{i}\label{eq8}\\
&x_{Pi}^{}=X_{Mi}^{}f/Z_{0},\;y_{Pi}^{}=Y_{Mi}^{}f/Z_{0}\label{eq9}
\end{align}
The theoretical development of POSIT is omitted, and its details can be acquired in \cite{15}. The pose of $M_0uvw$ relative to $Oxyz$ can be estimated by solving the following equations
\begin{equation}
\begin{split}
\begin{aligned}
\protect\overrightarrow{M_0M_i}\cdot\textbf{I}=\left(1+\varepsilon_{i}\right)x_{Mi}^{}-x_{M0}^{}
\end{aligned}
\end{split}
\label{14}\end{equation}
\begin{equation}
\begin{split}
\begin{aligned}
\protect\overrightarrow{M_0M_i}\cdot\textbf{J}=\left(1+\varepsilon_{i}\right)y_{Mi}^{}-y_{M0}^{}
\end{aligned}
\end{split}
\label{15}\end{equation}\vspace{-0.3em}where \begin{math}\textbf{I}=\frac{f}{Z_{0}}\cdot\vec{i}\end{math}, \begin{math}\varepsilon_{i}=\frac{1}{Z_{0}}\protect\overrightarrow{M_0M_i}\cdot \vec{k}\end{math}, \begin{math}\textbf{J}=\frac{f}{Z_{0}}\cdot\vec{j}\end{math}, and \begin{math}\vec{k}=\vec{i}\times \vec{j}\end{math}.

\subsection{Extended POSIT Algorithm}
Since the perspective projection model only fits for the pinhole camera, to generalize the algorithm to the fish-eye camera, the radially symmetric projection model \cite{25} is employed, which is shown in Fig. \ref{fig1:b}. In this figure, the meanings of the coordinate system $Oxyz$ and the symbol $f$ are the same as the definition in Fig. \ref{fig1:a}, and the image plane $G$ is represented by the coordinate system $Cx_Gy_G$. The object point $M$ is projected to $m$ for the pinhole camera, but for the fish-eye camera, due to the radial distortion it is projected to $m^{'}$. The incident angle between the line of sight for $M$ and the principal axis $z$ is $\theta$, the distance between the image point ($m$ or $m^{'}$) and the principal point $C$ is $r$, and $\varphi$ is the angle between $\overrightarrow{Cm}$ and $x_G$-axis. 
Based on the above descriptions, the projection model of pinhole camera can be expressed as
\vspace{-1em}
\begin{equation}
r_{1}=f\tan\theta \,\left(\text{\romannumeral1. perspective projection}\right)
\label{eq1}\end{equation}The projection model of fish-eye cameras can be described as follows
\vspace{-0.5em}
{\setlength{\abovecaptionskip}{1pt}
\setlength{\belowcaptionskip}{1pt}
\begin{align}
&r_{2}=2f\tan\left(\theta/2\right) \,\left(\text{\romannumeral2. stereographic projection}\right)\label{eq2}\\
&r_{3}=f\theta \,\left(\text{\romannumeral3. equidistance projection}\right)\label{eq3}\\
&r_{4}=2f\sin\left(\theta/2\right) \,\left(\text{\romannumeral4. equisolid projection}\right)\label{eq4}\\
&r_{5}=f\sin\theta \,\left(\text{\romannumeral5. orthogonal projection}\right)\label{eq5}
\end{align} }
In order to obtain a generic pose expression, the image point relationships between the pinhole camera and the fish-eye cameras are explored, and the results are shown in Table \ref{table1}.
\vspace{-0.5em}
\begin{table}[H]
\caption{The image point relationship between the pinhole camera and fish-eye cameras.}\label{table1}
 \setlength{\tabcolsep}{1.5mm}{
\begin{tabular}{c|ccccc}
\hline  
Projection model&\begin{math}r_{1}\end{math}&\begin{math}r_{2}\end{math}&\begin{math}r_{3}\end{math}&\begin{math}r_{4}\end{math}&\begin{math}r_{5}\end{math}\\
\hline  
Ratio \begin{math}\frac{{r}_{j}}{r_{1}}\end{math}&\begin{math}\frac{\cos\theta}{\cos\theta}\end{math}&\begin{math}\frac{\cos\theta}{\cos^{2}\left(\theta/2\right)}\end{math}&\begin{math}\frac{\cos\theta}{\left(\sin\theta/\theta\right)}\end{math}&\begin{math}\frac{\cos\theta}{\cos\left(\theta/2\right)}\end{math}&\begin{math}\frac{\cos\theta}{1}\end{math}\\
\hline 
\end{tabular}}\end{table}
\noindent Let the denominator of \begin{math}r_{j}/r_{1}\end{math} be $g_j(\theta)$, it can be obtained that
\begin{equation}g_j\left(\theta\right)\times \frac{r_{j}}{f}=\sin\theta, \ (j = 1,...5)\label{eq6}\end{equation} To get a concise expression, the ratio $r_{j}/r_{1}$ is denoted as $G_j\left(\theta\right)$, i.e. \begin{math}G_j\left(\theta\right)=\cos\theta/g_j\left(\theta\right)\end{math}. 

According to $G_j\left(\theta\right)$ and (\ref{eq7})-(\ref{eq9}), the image coordinates obtained by different cameras can be expressed as
\begin{align}
&x_{M0^{'}}^{}=x_{M0}^{}G_j\left(\theta_{M0}^{}\right),\ y_{M0^{'}}^{}=y_{M0}^{}G_j\left(\theta_{M0}^{}\right)\label{eq10}\\
&x_{Mi^{'}}^{}=x_{Mi}^{}G_j\left(\theta_{Mi}^{}\right),\ y_{Mi^{'}}^{}=y_{Mi}^{}G_j\left(\theta_{Mi}^{}\right)\label{eq11}\\
&x_{Pi^{'}}^{}=x_{Pi}^{}G_j\left(\theta_{Pi}^{}\right),\ y_{Pi^{'}}^{}=y_{Pi}^{}G_j\left(\theta_{Pi}^{}\right)\label{eq12}
\end{align}where $\theta_{M0}^{}$, $\theta_{Mi}^{}$ and $\theta_{Pi}^{}$ respectively represent the incident angles of point $M_0$, $M_i$ and $P_i$. 

Due to page limitation, a conclusion is given directly without proof$\footnote{The proof is available at https://github.com/k032131/EPOSIT.}$.

\textbf{Conclusion}: Without loss of generality, suppose \begin{math}0\leq \theta_{M0}^{}\leq \theta_{Pi}^{}<90^{\circ}\end{math}. For the fish-eye cameras, if \begin{math}\left|\protect\overrightarrow{M_0M_i}\right| \leq 0.1Z_{0}\end{math}, \begin{math}G_j\left(\theta_{M0}^{}\right)\end{math} and \begin{math}G_j\left(\theta_{Pi}^{}\right)\end{math} can be regarded as the same i.e. \begin{math}G_j\left(\theta_{M0}^{}\right)\approx G_j\left(\theta_{Pi}^{}\right)\end{math}.

Based on this conclusion, the expression of (\ref{eq12}) can be rewritten as 
{\setlength\abovedisplayskip{1pt}
\setlength\belowdisplayskip{1pt}
\begin{equation}
\begin{split}
\begin{aligned}
x_{Pi^{'}}^{}&=\frac{f}{Z_{0}}\left(X_{Mi}^{}G\left(\theta_{Pi}^{}\right)-X_{M0}^{}G\left(\theta_{M0}^{}\right)\right)+x_{M0^{'}}^{}\\
&\approx x_{M0^{'}}+\frac{f}{Z_{0}}G\left(\theta_{M0}^{}\right)\left(X_{Mi}^{}-X_{M0}^{}\right)\\
&=x_{M0^{'}}^{}+s\left(X_{Mi}^{}-X_{M0}^{}\right)
\end{aligned}
\end{split}
\label{eq13}\end{equation}}
{\setlength\abovedisplayskip{1pt}
\setlength\belowdisplayskip{1pt}
\begin{equation}
\begin{split}
\begin{aligned}
y_{Pi^{'}}^{}&=\frac{f}{Z_{0}}\left(Y_{Mi}^{}G\left(\theta_{Pi}^{}\right)-Y_{M0}^{}G\left(\theta_{M0}^{}\right)\right)+y_{M0^{'}}^{}\\
&\approx y_{M0^{'}}+\frac{f}{Z_{0}}G\left(\theta_{M0}^{}\right)\left(Y_{Mi}^{}-Y_{M0}^{}\right)\\
&=y_{M0^{'}}^{}+s\left(Y_{Mi}^{}-Y_{M0}^{}\right)
\end{aligned}
\end{split}
\label{eq14}\end{equation}}\vspace{0.5em}where $s=\frac{f}{Z_0}G\left(\theta_{M0}\right)$ is the scaling factor of the SOP. From Fig. \ref{fig1:a}, it can be seen that
\begin{equation}\protect\overrightarrow{M_0M_i}=\protect\overrightarrow{M_0N_i}+\protect\overrightarrow{N_iP_i}+\protect\overrightarrow{P_iM_i}\label{eq15}\end{equation}In perspective projection model, points $M_i$ and $N_i$ are projected at the same image point, therefore, (\ref{eq11}) can be rewritten as
{\setlength\abovedisplayskip{5pt}
\setlength\belowdisplayskip{5pt}
\begin{equation}\left(x_{Mi^{'}}^{},y_{Mi^{'}}^{}\right)^T=f\frac{\left(X_{Ni}^{},Y_{Ni}^{}\right)^T}{Z_{0}}G\left(\theta_{Mi}^{}\right)\label{eq16}\end{equation}}where \begin{math}\left(X_{Ni}^{},Y_{Ni}^{}\right)^T\end{math} represents the \begin{math}x\end{math} and \begin{math}y\end{math} coordinates of point \begin{math}N_{i}\end{math} in the camera coordinate system. According to (\ref{eq7}), (\ref{eq10}) and (\ref{eq16}), $\protect\overrightarrow{M_0N_i}$ can be expressed as
\begin{equation}
\begin{split}
\begin{aligned}
\protect\overrightarrow{M_0N_i}&=\left(X_{Ni}^{},Y_{Ni}^{}\right)^T-\left(X_{M0}^{},Y_{M0}^{}\right)^T\\
&=\frac{Z_{0}}{f}\textstyle\left(\frac{\left(x_{Mi^{'}}^{},y_{Mi^{'}}^{}\right)^T}{G\left(\theta_{Mi}^{}\right)}-\frac{\left(x_{M0^{'}}^{},y_{M0^{'}}^{}\right)^T}{G\left(\theta_{M0}^{}\right)}\right)
\end{aligned}
\end{split}
\label{eq17}\end{equation}Since triangle \begin{math}\bigtriangleup Cm_{i}O\end{math} is similar to triangle \begin{math}\bigtriangleup P_{i}N_{i}M_{i}\end{math}, it is obtained that
\begin{equation}\textstyle\frac{\left|\protect\overrightarrow{P_iM_i}\right|}{\left|\protect\overrightarrow{OC}\right|}=\frac{\protect\overrightarrow{M_0M_i}\cdot \vec{k}}{f}=\frac{\left|\protect\overrightarrow{N_iP_i}\right|}{\left|\protect\overrightarrow{Cm_i}\right|}\label{eq18}\end{equation}Substituting (\ref{eq11}) into (\ref{eq18}), $\protect\overrightarrow{N_iP_i}$ can be expressed as
\begin{equation}\protect\overrightarrow{N_iP_i}=\textstyle\frac{\protect\overrightarrow{M_0M_i}\cdot \vec{k}}{f}\cdot \frac{\protect\overrightarrow{Cm_{i^{'}}}}{G\left(\theta_{Mi}^{}\right)}\label{eq19}\end{equation}Substituting (\ref{eq17}) and (\ref{eq19}) into (\ref{eq15}), $\protect\overrightarrow{M_{0}M_{i}}$ equals to
\begin{equation}
\begin{split}
\begin{aligned}
\protect\overrightarrow{M_{0}M_{i}}&=\frac{Z_{0}}{f}\textstyle\left(\frac{\left(x_{Mi^{'}}^{},y_{Mi^{'}}^{}\right)^T}{G\left(\theta_{Mi}^{}\right)}-\frac{\left(x_{M0^{'}}^{},y_{M0^{'}}^{}\right)^T}{G\left(\theta_{M0}^{}\right)}\right)\\
&+\textstyle\frac{\protect\overrightarrow{M_{0}M_{i}}\cdot \vec{k}}{f}\cdot \frac{\protect\overrightarrow{Cm_{i^{'}}}}{G\left(\theta_{Mi}^{}\right)}+\protect\overrightarrow{P_{i}M_{i}}
\end{aligned}
\end{split}
\label{eq20}\end{equation}From Fig. \ref{fig1:a}, \begin{math}\protect\overrightarrow{P_{i}M_{i}}\cdot\vec{i}=0\end{math} and \begin{math}\protect\overrightarrow{P_{i}M_{i}}\cdot\vec{j}=0\end{math}. Multiplied by $\frac{f}{Z_{0}}\cdot\vec{i}$ and $\frac{f}{Z_{0}}\cdot\vec{j}$ respectively on both sides of (\ref{eq20}), it can be obtained that
\begin{equation}
\begin{split}
\begin{aligned}
\frac{f}{Z_{0}}\protect\overrightarrow{M_{0}M_{i}}\cdot\vec{i}=\frac{1+\varepsilon_{i}}{G\left(\theta_{Mi}^{}\right)}x_{Mi^{'}}^{}-\frac{1}{G\left(\theta_{M0}^{}\right)}x_{M0^{'}}^{}
\end{aligned}
\end{split}
\label{eq21}\end{equation}
\begin{equation}
\begin{split}
\begin{aligned}
\frac{f}{Z_{0}}\protect\overrightarrow{M_{0}M_{i}}\cdot\vec{j}=\frac{1+\varepsilon_{i}}{G\left(\theta_{Mi}^{}\right)}y_{Mi^{'}}^{}-\frac{1}{G\left(\theta_{M0}^{}\right)}y_{M0^{'}}^{}
\end{aligned}
\end{split}
\label{eq22}\end{equation}According to the definition of $\textbf{I}$ and $\textbf{J}$ in (\ref{14}) and (\ref{15}), we have
\begin{equation}
\protect\overrightarrow{M_{0}M_{i}}\cdot\textbf{I} =\xi_{i}\label{add21}
\end{equation}
\begin{equation}
\protect\overrightarrow{M_{0}M_{i}}\cdot\textbf{J}=\eta_{i}\label{add22}
\end{equation}where $\xi_{i}$ and $\eta_{i}$ represent the right-hand sides of equations (\ref{eq21}) and (\ref{eq22}) respectively.

For a given image point, its corresponding incident angle $\theta$ can be obtained according to (\ref{eq1})-(\ref{eq5}). To calculate $\theta$, the principal point $C$ is needed. Similar to some PnPfr methods, such as the algorithms in \cite{16}, \cite{19} and \cite{21}, the image center is regarded as the principal point. It will be shown in simulation and experiment that this assumption yields good results even though it is not strictly true. $G\left(\theta_{M0}\right)$ and $G\left(\theta_{Mi}\right)$ can be obtained based on the incident angle of image points $M_0$ and $M_i$. Therefore, only $\varepsilon_{i}$ is unknown in $\xi_{i}$ and $\eta_{i}$.
Given four object points and their corresponding image points, according to (\ref{add21}) and (\ref{add22}), a linear system for the coordinates of the unknown vectors $\textbf{I}$ and $\textbf{J}$ can be described as 
{\setlength\abovedisplayskip{1pt}
\setlength\belowdisplayskip{1pt}
\begin{align}
&\textbf{A}\textbf{I}=x^{'}\label{eq24}\\
&\textbf{A}\textbf{J}=y^{'}\label{eq25}
\end{align}}where \begin{math}\textbf{A}\end{math} is the matrix of the coordinates of \begin{math}\protect\overrightarrow{M_0M_{i}}\end{math} in the object coordinate system, \begin{math}x^{'}\end{math} is the vector with the \begin{math}i\end{math}-th coordinate \begin{math}\xi_{i}\end{math} and \begin{math}y^{'}\end{math} is the vector with the \begin{math}i\end{math}-th coordinate \begin{math}\eta_{i}\end{math}. If these four points are noncoplanar, matrix $\textbf{A}$ is invertible. According to (\ref{eq24}) and (\ref{eq25}), accurate pose results can be estimated by some iterative methods, such as the one proposed in \cite{15}.

Comparing (\ref{14})-(\ref{15}) with (\ref{eq21})-(\ref{eq22}), it can be seen that the expression of EPOSIT is similar to the POSIT algorithm. This means that the good properties of the POSIT algorithm are also reserved in EPOSIT, such as high efficiency and unique solution. Since $G_j(\theta)$ is different for  each camera model, it seems that the specific camera model is required in advance. However, in simulation, we will show that the estimation results of different fish-eye cameras can be regarded as the same. Therefore, only the pinhole camera or the fish-eye camera need to be distinguished for users, which is simple in practical applications.

\begin{figure*}
    \centering
     \subfigure[]
    {
        \includegraphics[width=2.2in]{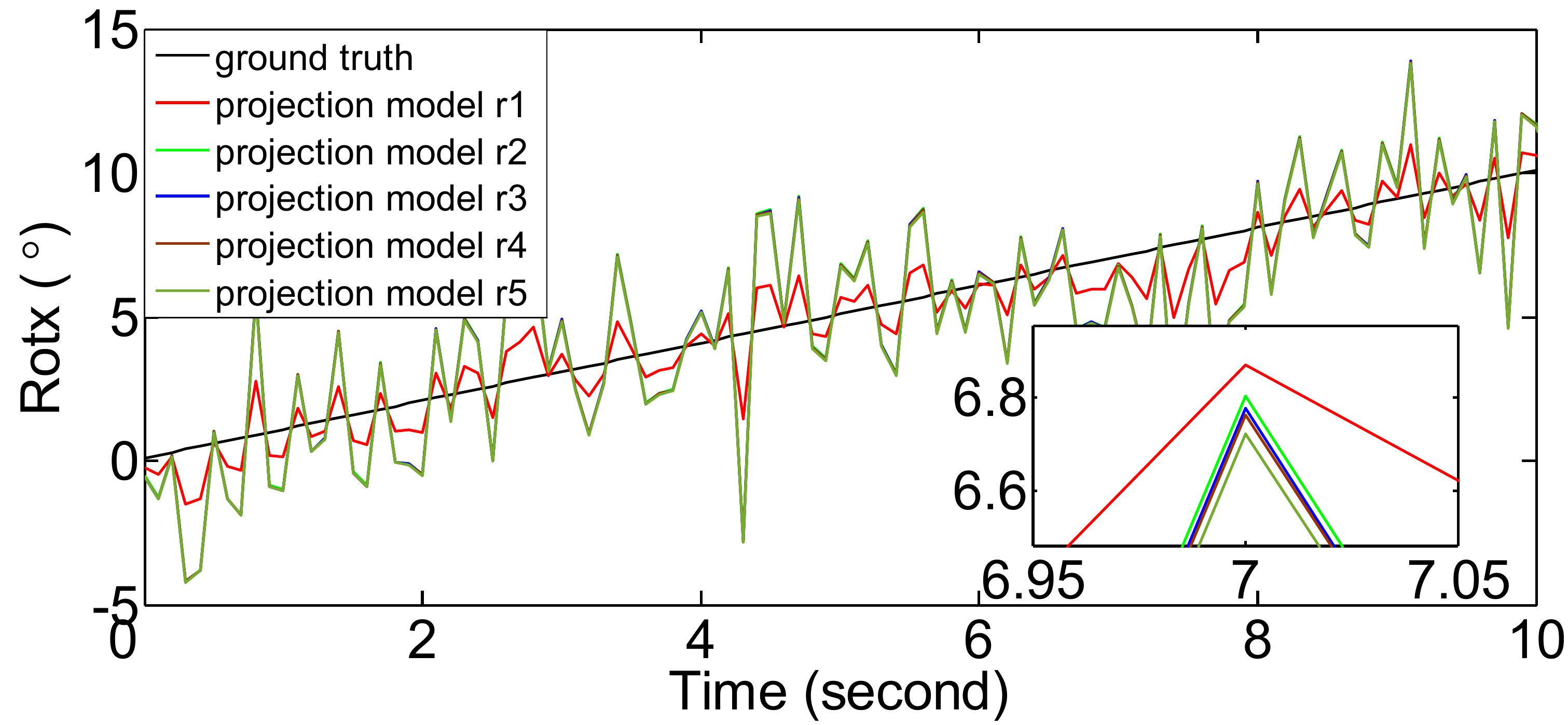} \label{20190902fig6:a}
    }
\vspace{-0.3em}
    \subfigure[]
    {
        \includegraphics[width=2.2in]{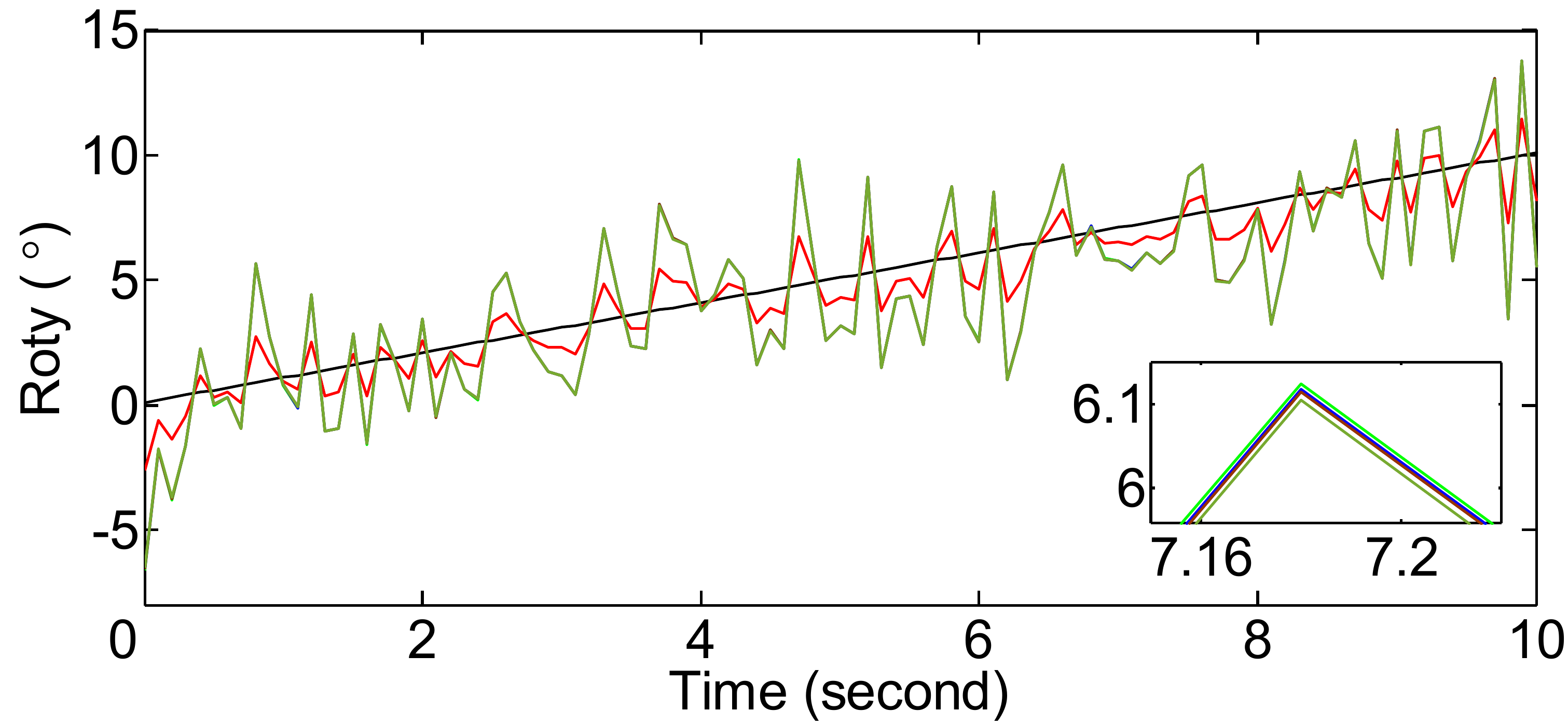}\label{20190902fig6:b}
    }
\vspace{-0.3em}
    \subfigure[]
    {
        \includegraphics[width=2.2in]{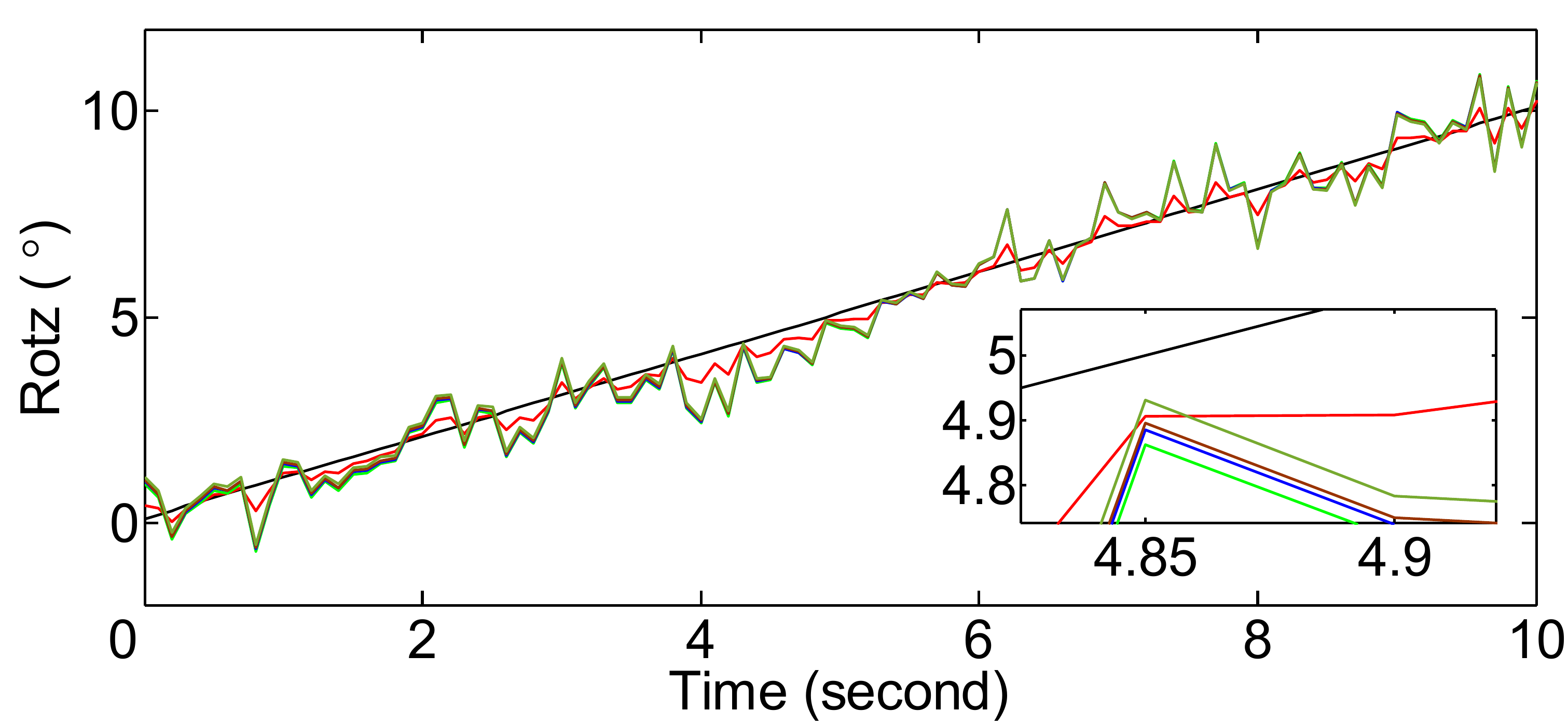} \label{20190902fig6:c}
    }
\vspace{-0.3em}
    \\
    \subfigure[]
    {
        \includegraphics[width=2.2in]{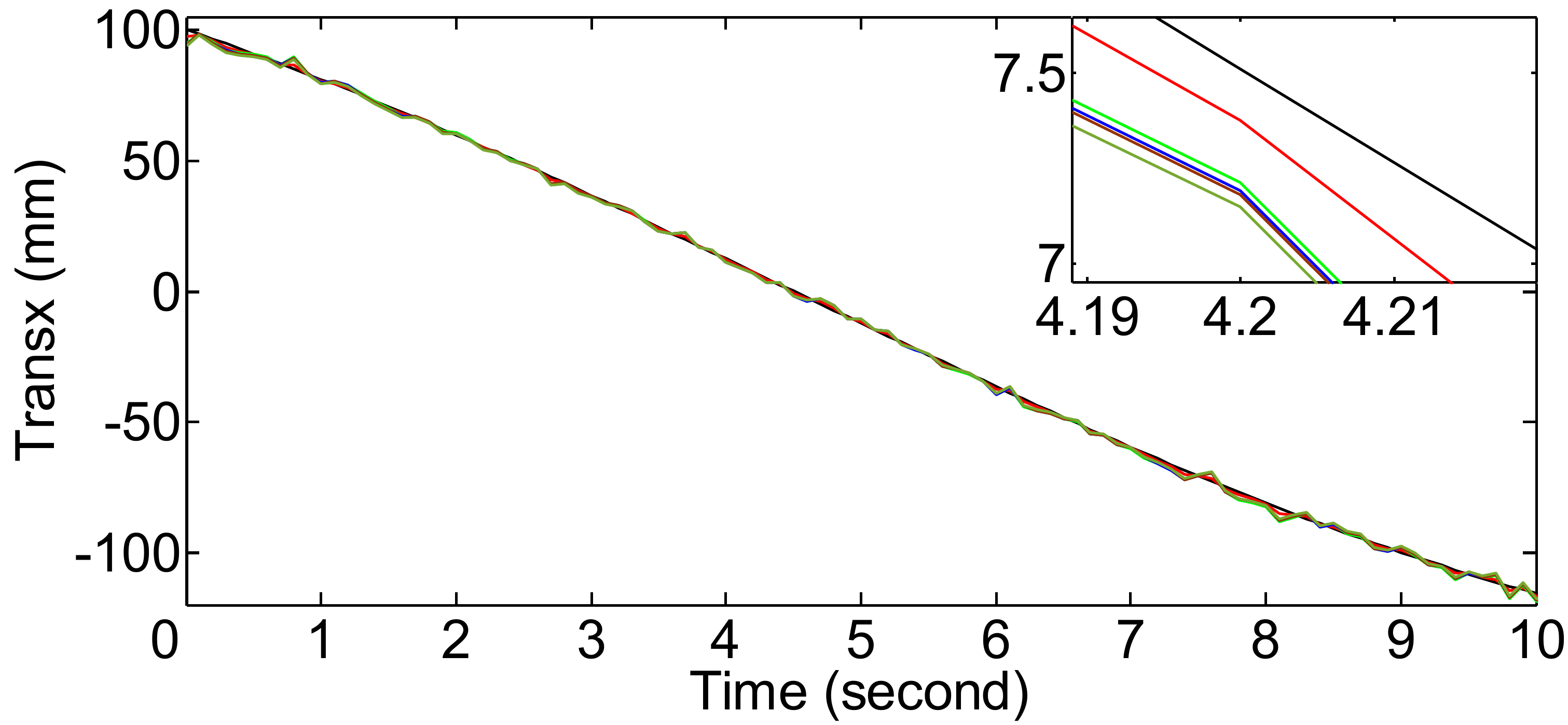}\label{20190902fig6:d}
    }
    \subfigure[]
    {
        \includegraphics[width=2.2in]{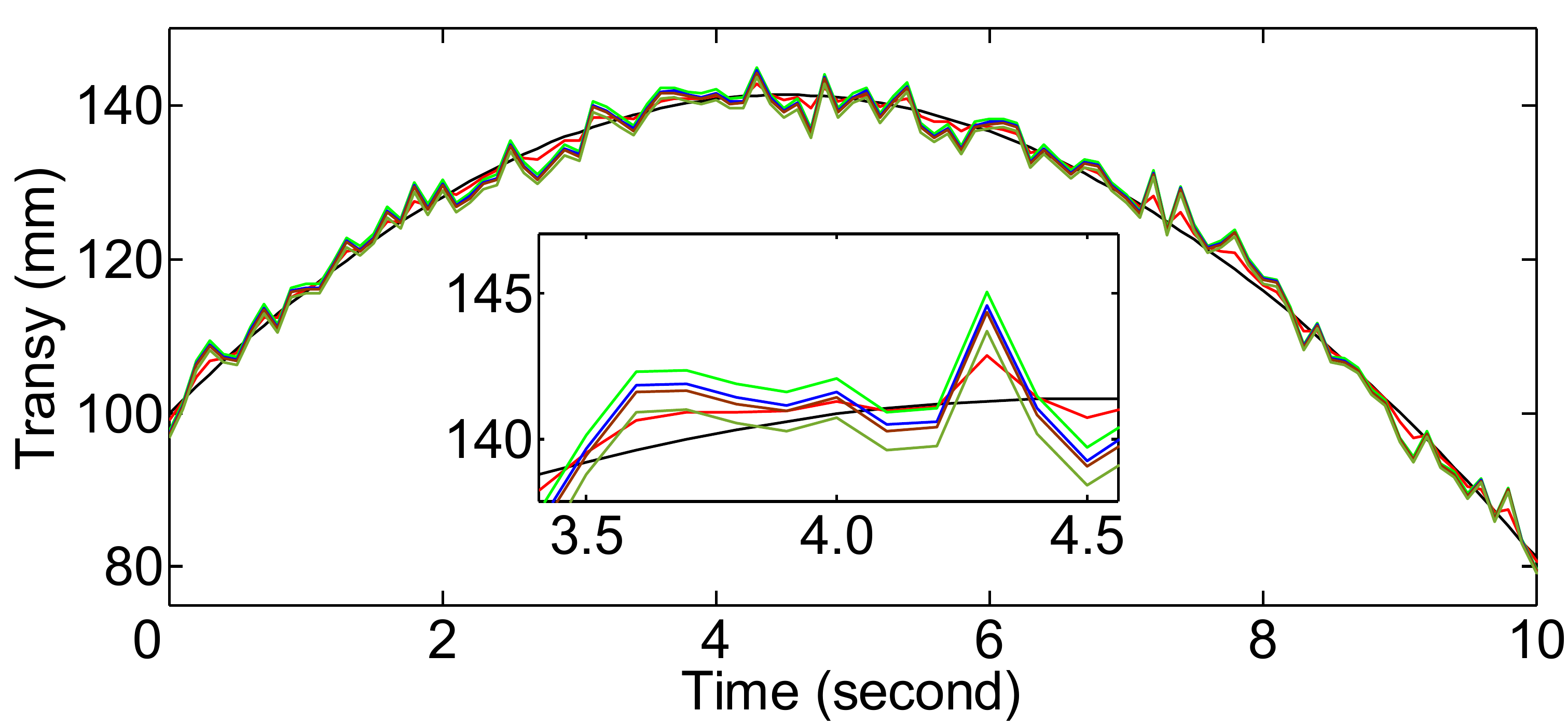} \label{20190902fig6:e}
    }
    \subfigure[]
    {
        \includegraphics[width=2.2in]{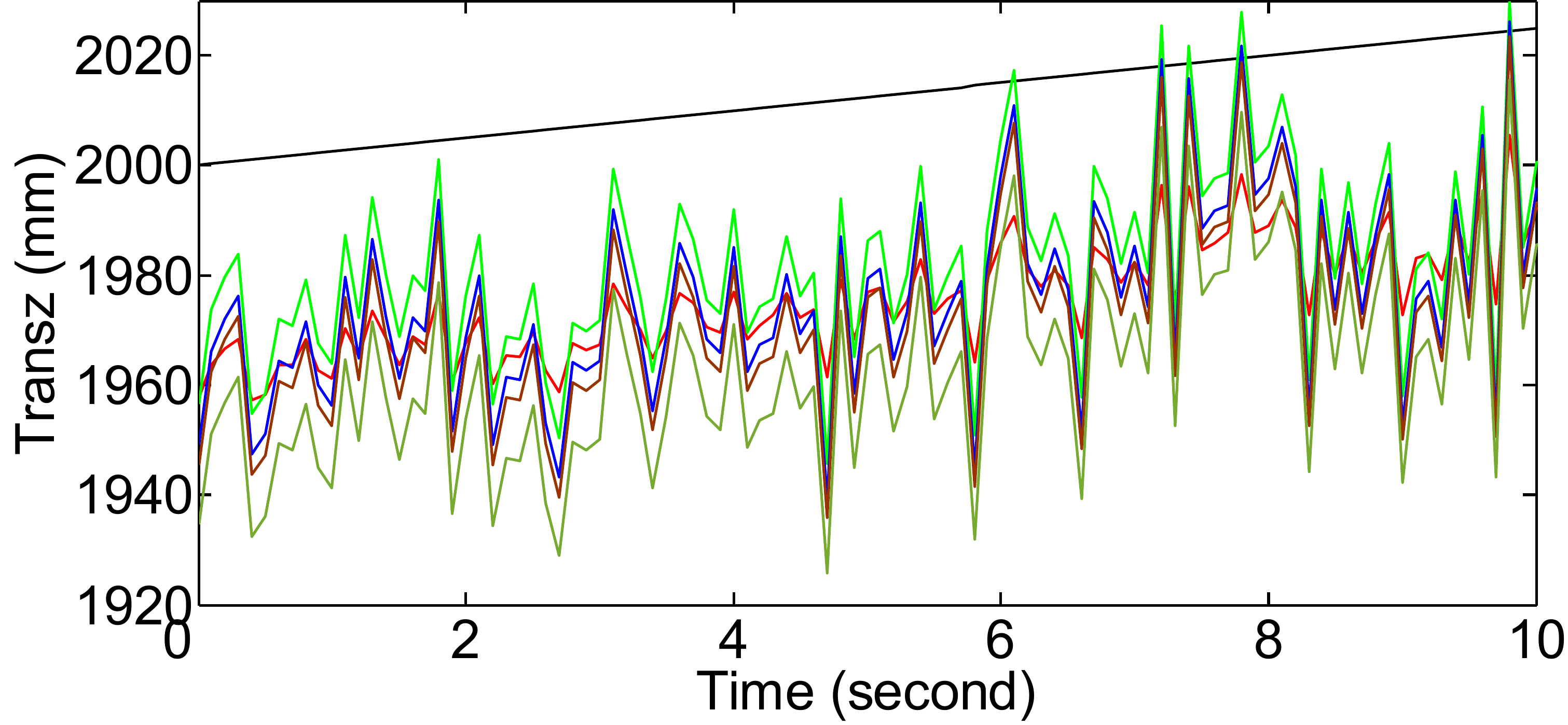}\label{20190902fig6:f}
    }
    \caption{Simulation results. Top row: truth values and estimation results of rotation around $x$, $y$ and $z$ axes respectively; bottom row: truth values and estimation results of translation in the $x$, $y$ and $z$ directions respectively.} 
\label{20190902fig6} 
\end{figure*}
\begin{figure}
    \centering
     \subfigure[]
    {
        \includegraphics[width=3.2in]{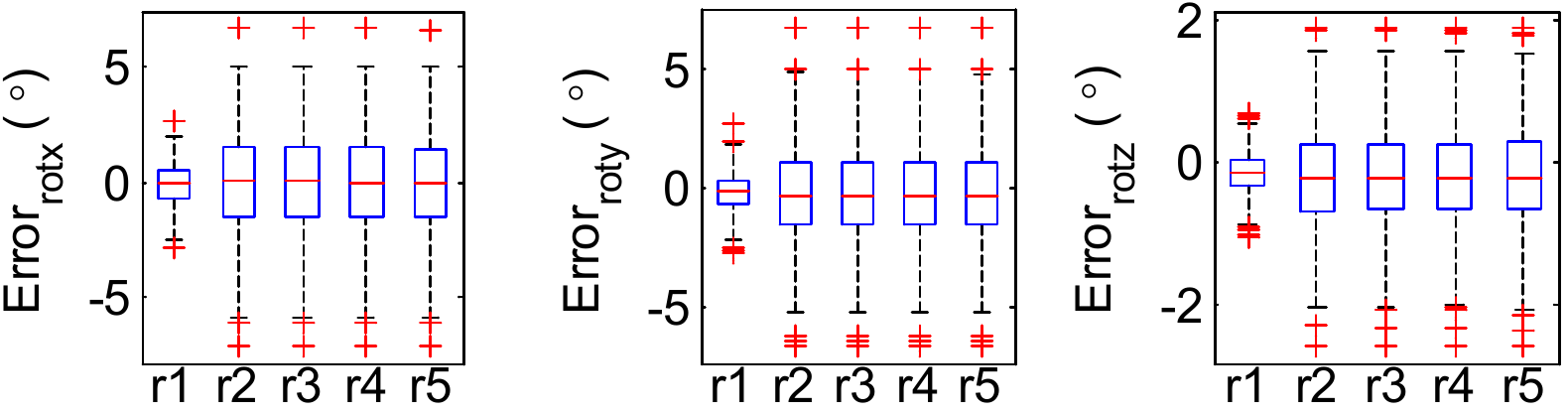} \label{fig7:b}
    }
    \subfigure[]
    {
        \includegraphics[width=3.2in]{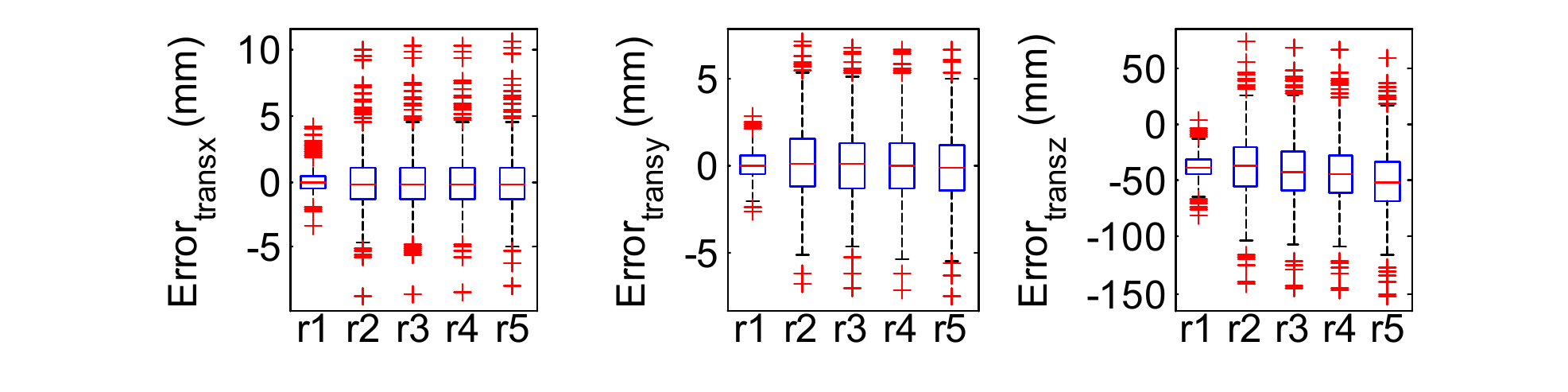}\label{fig7:c}
    }
    \caption{Boxplot of estimation errors. Top row: estimation errors of the rotation; bottom row: estimation errors of the translation.} 
    \label{20190823fig1}
\end{figure}

\textbf{Remark 1}: Although the focal length is required in EPOSIT compared with PnPfr methods, we argue that this condition is easy to be satisfied, because the focal length used in this paper can be obtained from instructions for use.

\section{Simulation results}
In simulation, the precision of EPOSIT algorithm and its stability to the principal point error are shown. Another three state-of-the-art solvers, OPnP \cite{33}, GPnPf \cite{17}, and PnPfr \cite{21} are compared. OPnP is a globally optimal PnP solver, where the camera pose can be calculated based on accurate camera intrinsic parameters and four 3D-2D point correspondences. GPnPf can determine the camera pose with unknown focal length using four 3D-2D point correspondences. PnPfr proposes a minimal solver to P4P problems with unknown focal length and radial distortion.

Four noncoplanar object points, whose coordinates in the object coordinate system are ($0,0,0$), ($200mm,0,0$), ($0,200mm,0$), ($0,0,-50mm$), and their corresponding image points are employed in these methods. The focal lengths of the fish-eye camera and the pinhole camera are 541 pixels and 1353 pixels, respectively. The intrinsic parameters are \begin{math}k_{x}\end{math}=552.39, \begin{math}k_{y}\end{math}=552.69, \begin{math}u_{0}\end{math}=782.41, \begin{math}v_{0}\end{math}=613.71 and \begin{math}k_{x}\end{math}=1378.65, \begin{math}k_{y}\end{math}=1381.37, \begin{math}u_{0}\end{math}=807.41, \begin{math}v_{0}\end{math}=602.47, respectively. The Gaussian noise with the mean value 0 and the standard deviation 0.4 is added to image points. The object coordinate system rotates around the fixed camera coordinate system with the angular velocity \begin{math}0.1^{\circ}/s \end{math} and translates following a cylindrical helix which is expressed as 
\begin{align*}
&X(t)=100\sqrt{2}\cos(45^{\circ}+t)\\
&Y(t)=100\sqrt{2}\sin(45^{\circ}+t)\\
&Z(t)=2000+0.25t
\end{align*}

\subsection{Precision of EPOSIT}

The camera pose estimation results of EPOSIT are shown in Fig. \ref{20190902fig6}, and their corresponding errors are depicted in Fig. \ref{20190823fig1}.
In Fig. \ref{20190823fig1}, the mean values of translation errors and rotation errors are all nearly zero for different cameras except the mean values of translation error in the $z$ direction, which are about 50$mm$. Because the smallest truth value in the $z$ direction is 2000$mm$, 50$mm$ is acceptable in most practical applications.
According to Fig. \ref{20190902fig6} and Fig. \ref{20190823fig1}, the performances of these four fish-eye cameras are nearly the same. Comparing their results, the maximum difference of the mean value is 0.01$^\circ$ for the rotation error, 0.02$mm$ for the translation error in $x$ and $y$ directions, and 6$mm$ for the $z$ direction. Therefore, the projection model of the fish-eye camera is not necessary to be predetermined for users, which makes EPOSIT easy to be used.

In summary, EPOSIT can obtain precise estimation results for both the pinhole camera and the fish-eye camera.
\begin{figure}
    \centering
     \subfigure[]
    {
        \includegraphics[width=3.2in]{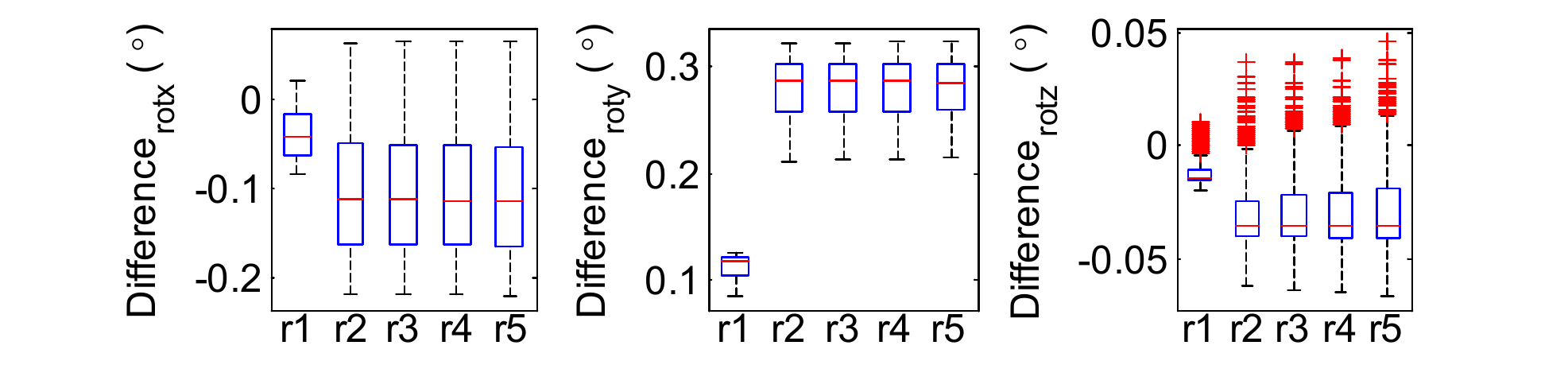} \label{20190823fig2:a}
    }
    \subfigure[]
    {
        \includegraphics[width=3.2in]{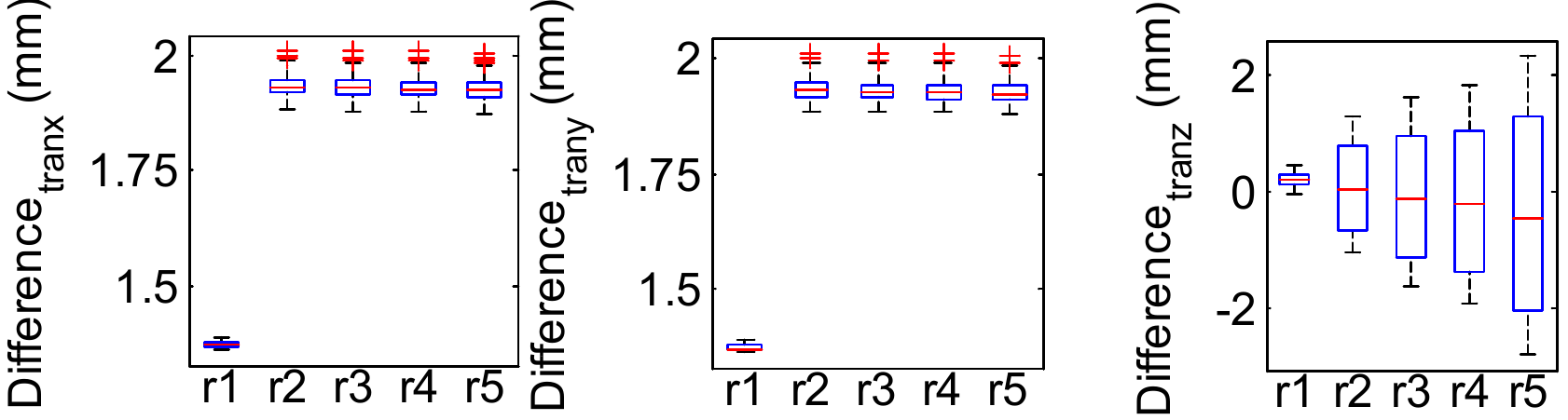}\label{20190823fig2:b}
    }
    \caption{Stability of EPOSIT to the principal point error. Top row: differences of rotation estimation results acquired by the principal point and the image center; bottom row: differences of translation estimation  results acquired by the principal point and the image center. } 
    \label{20190823fig2}
\end{figure}

\subsection{Stability of EPOSIT to the Principal Point Error}
In Section II-B, we have mentioned that the principal point is approximated by the image center. In order to demonstrate the feasibility of this approximation, the differences of estimation results respectively acquired by the precise principal point and the image center are shown in Fig. \ref{20190823fig2}. Note that the differences of the pinhole camera are smaller than the values of fish-eye cameras. The reason is $G(\theta)=1$ for the pinhole camera, hence the estimation results are less affected by the error of incident angle caused by the principal point error. From Fig. \ref{20190823fig2:a}, it can be seen that the largest mean value of differences of three axes is less than 0.3$^\circ$ and their deviations are also small. From Fig. \ref{20190823fig2:b}, the largest difference value is less than 1.9$mm$. It is acceptable comparing with the smallest truth value 100$mm$.

In conclusion, the differences of estimation results acquired by the precise principal point and the image center  are small and stable. The principal point can be substituted by the image center.

\subsection{Comparison  Results for the Pinhole Camera}
\begin{figure}[htb]
    \centering
     \subfigure[]
    {
        \includegraphics[width=1.58in]{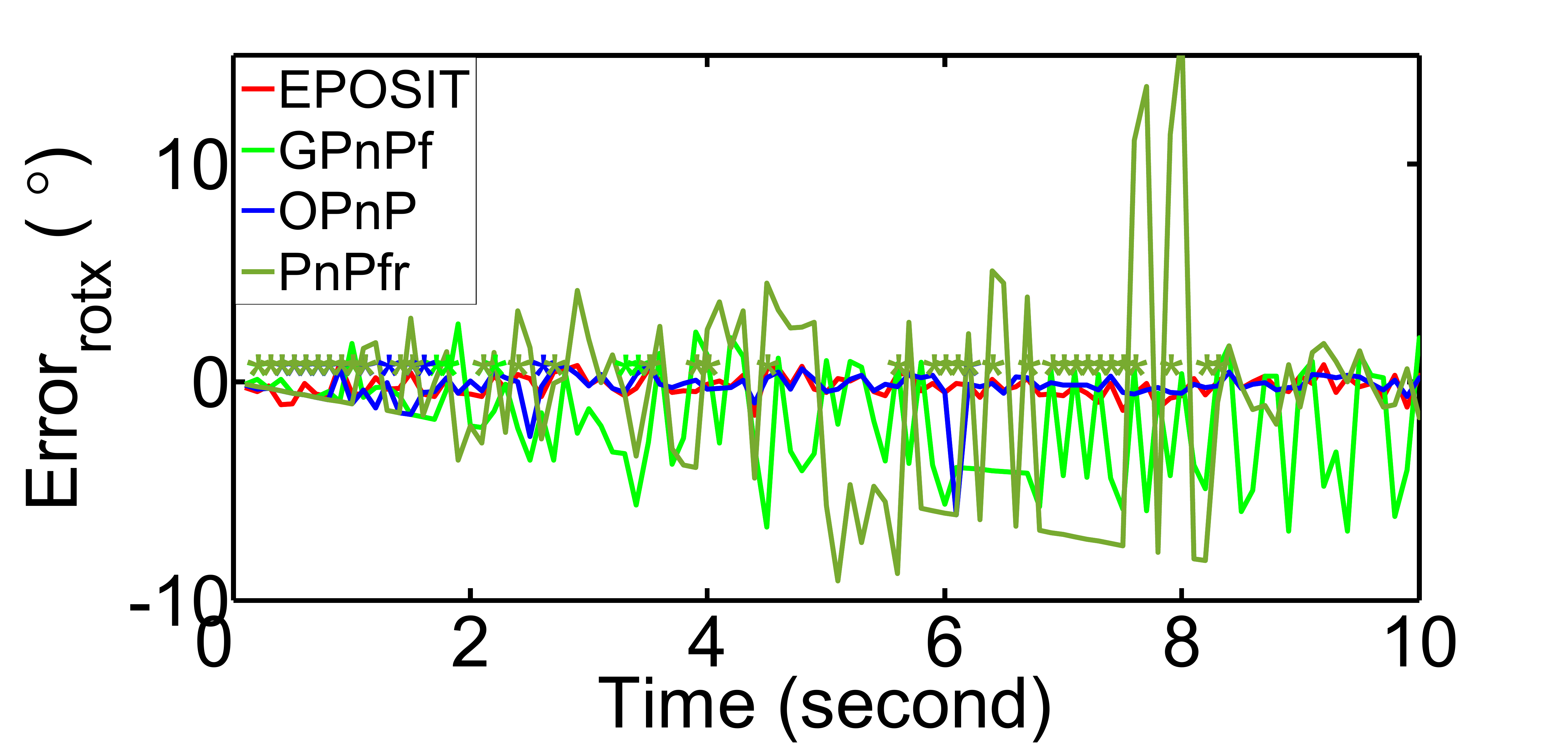} \label{fig8:a}
    }
\vspace{-0.45em}
    \subfigure[]
    {
        \includegraphics[width=1.58in]{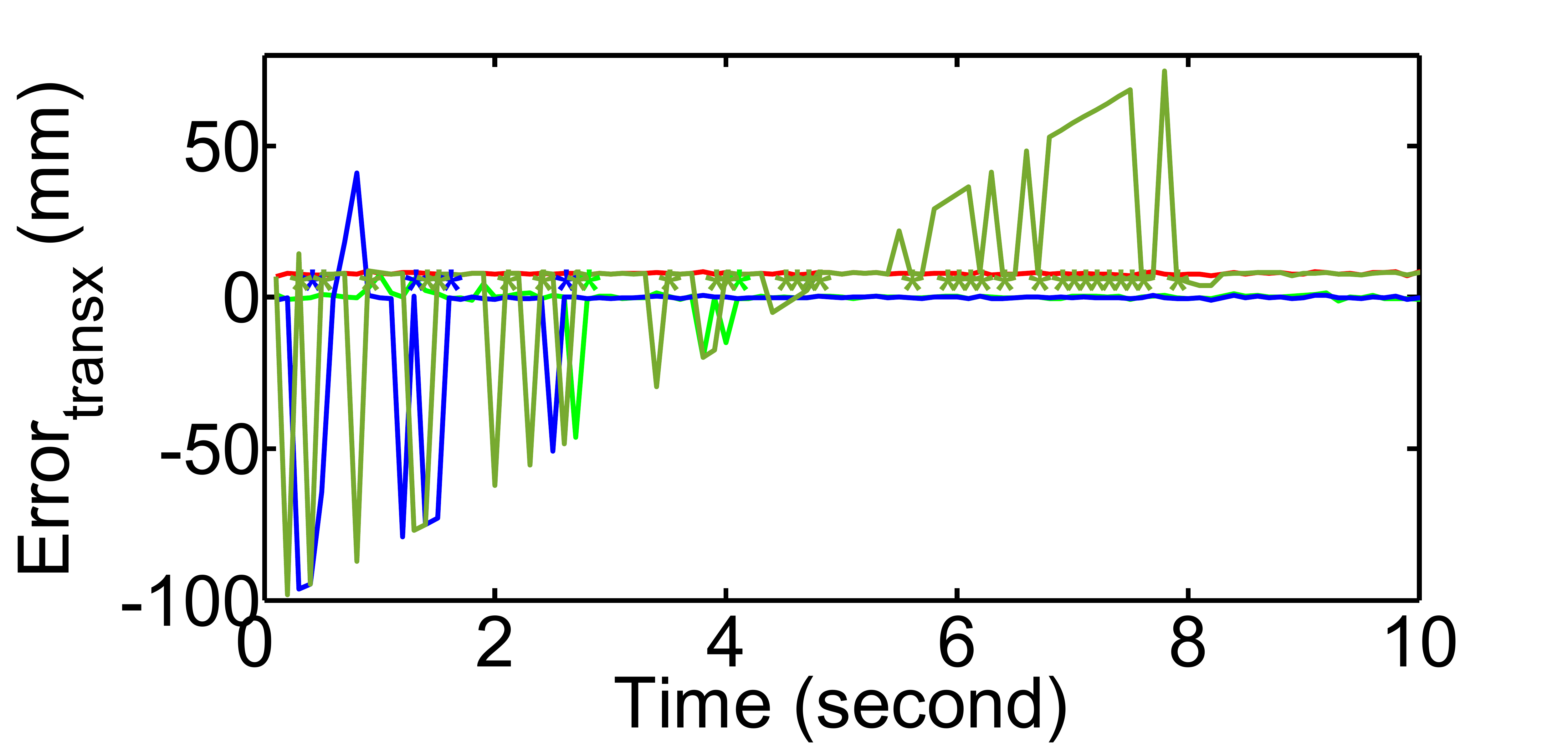}\label{fig8:b}
    }
\vspace{-0.45em}
    \\
    \subfigure[]
    {
        \includegraphics[width=1.58in]{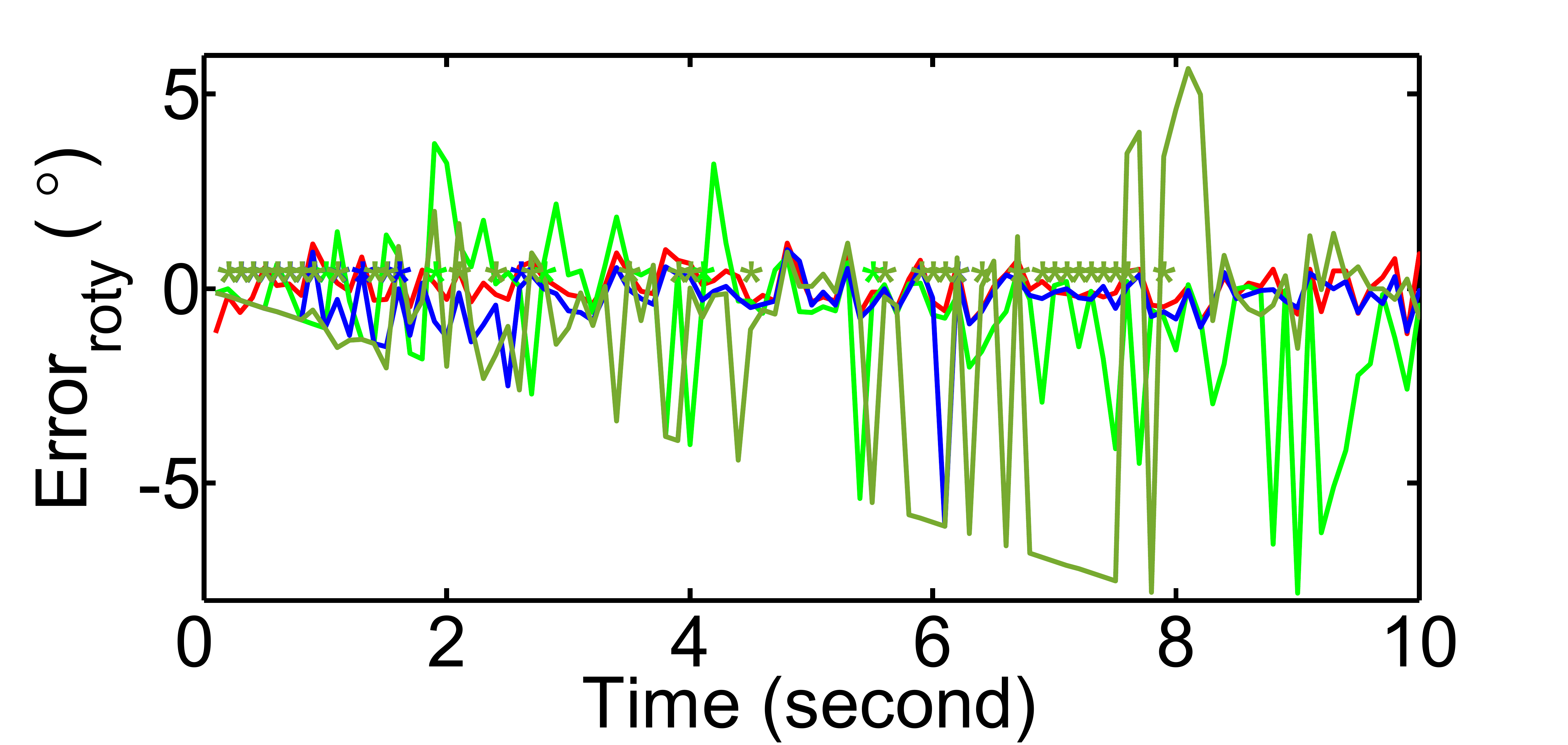} \label{fig8:c}
    }
\vspace{-0.45em}
    \subfigure[]
    {
        \includegraphics[width=1.57in]{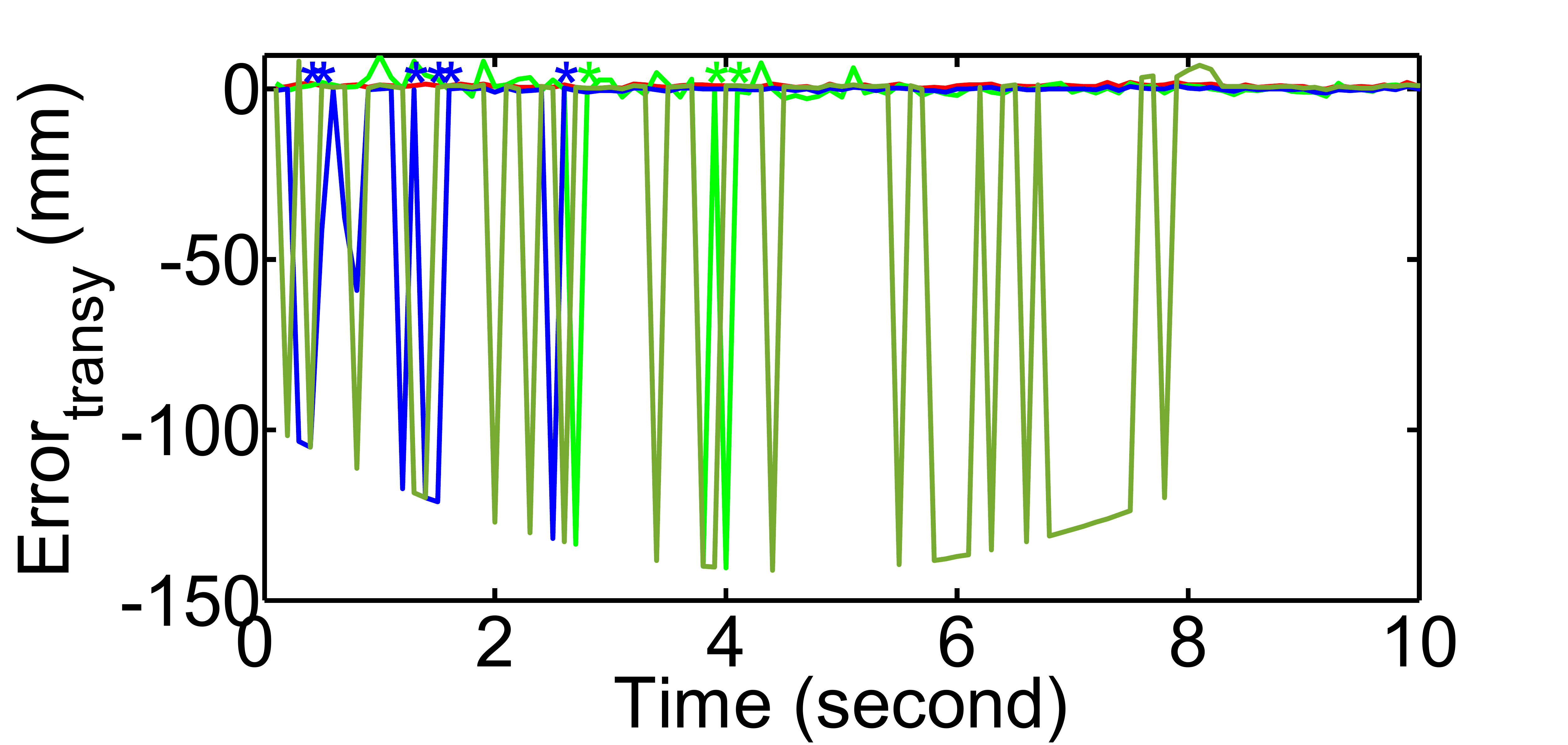}\label{fig8:d}
    }
\vspace{-0.45em}
 \\
    \subfigure[]
    {
        \includegraphics[width=1.58in]{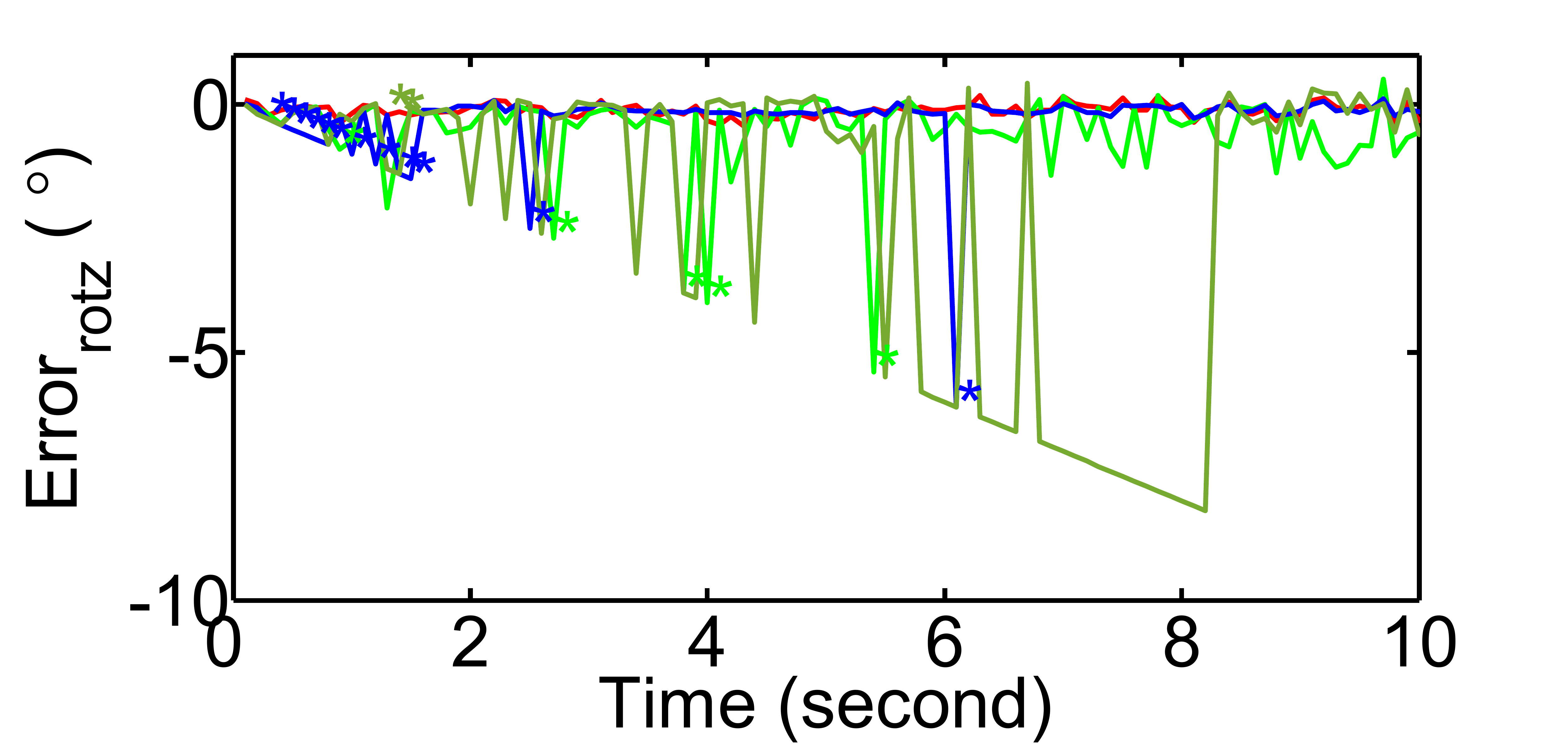} \label{fig8:e}
    }
    \subfigure[]
    {
        \includegraphics[width=1.58in]{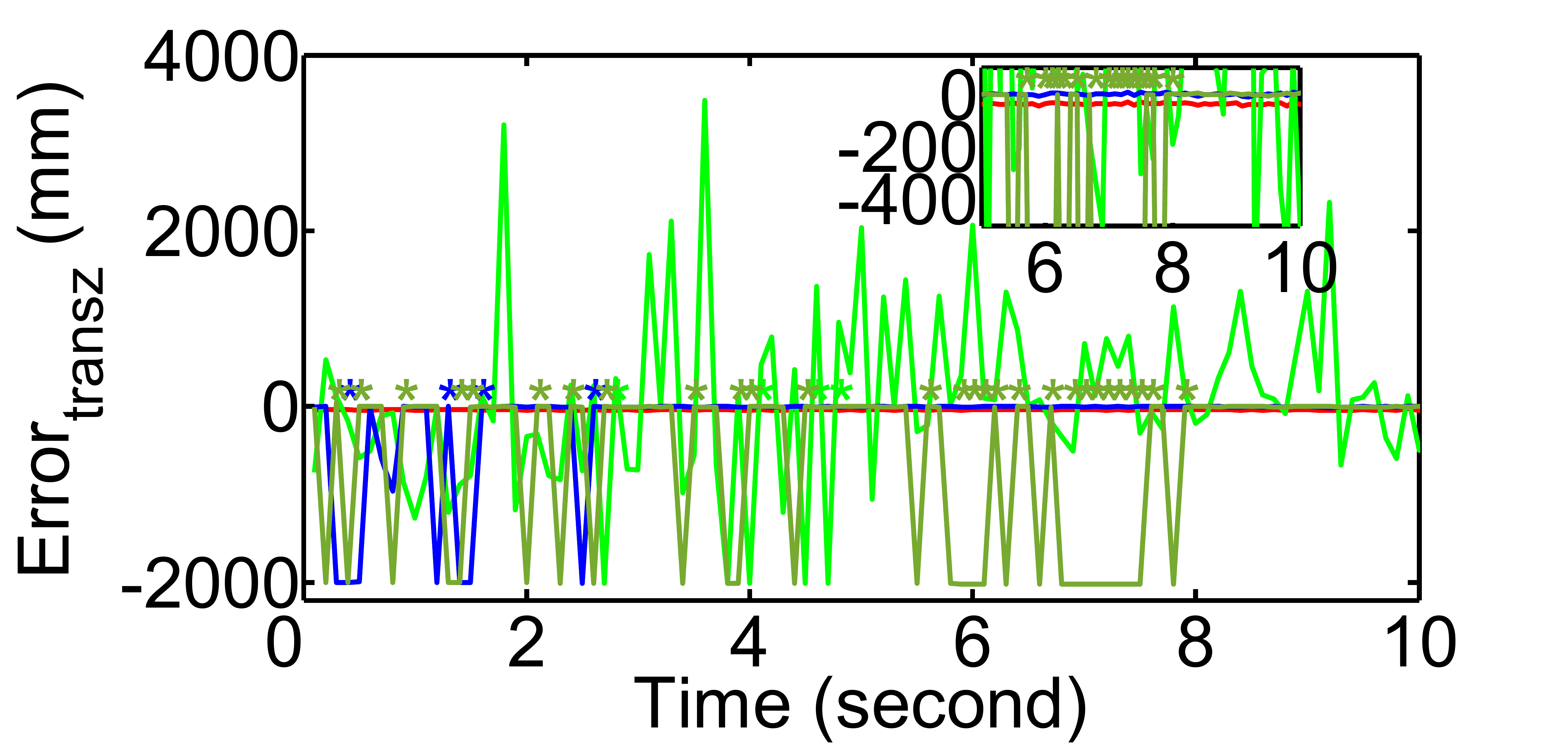}\label{fig8:f}
    }
   \caption{Comparison results for the pinhole camera. The left side represents estimation errors of the rotation, and the right side represents estimation errors of the translation.} 
    \label{20190902fig1}
\end{figure}

To further demonstrate the effectiveness of EPOSIT for the pinhole camera, three PnP solvers, GPnPf, OPnP and PnPfr are compared. Since for the pinhole camera, EPOSIT is degenerated to the POSIT algorithm, the comparison results with POSIT are not given. 
For EPOSIT, GPnPf and PnPfr, the image center is regarded as the principal point, and accurate camera intrinsic parameters are employed for OPnP. Other parameters of the compared methods are the same as their original literatures. The estimation errors of these algorithm are shown in Fig. \ref{20190902fig1}. In this figure, the symbol $\star$ represents the estimation result at least twice larger than the ground truth or the solver cannot obtain the solution. From Fig. \ref{20190902fig1}, it is clear that EPOSIT performs the best compared with other methods. The results of OPnP are similar to EPOSIT, but at some points it is worse than ours. The reason why the performance of OPnP is better than GPnPf and PnPfr is that accurate calibration results are employed in OPnP, therefore, the number of unknown variables in OPnP is smaller than the other two methods. OPnP also has some abnormal solutions due to image noise, which demonstrates the robustness of EPOSIT. According to comparison  results, although the focal length is not required in GPnPf and PnPfr, the accuracies of them are less than EPOSIT. Moreover, the focal length employed in EPOSIT is easy to be acquired from instructions for use. Therefore, our research is still meaningful.  

In summary, EPOSIT can achieve a good performance in precision and stability for the pinhole camera pose estimation. 

\subsection{Comparison Results for the Fish-Eye Camera}
\begin{figure}[htb]
    \centering
     \subfigure[]
    {
        \includegraphics[width=1.58in]{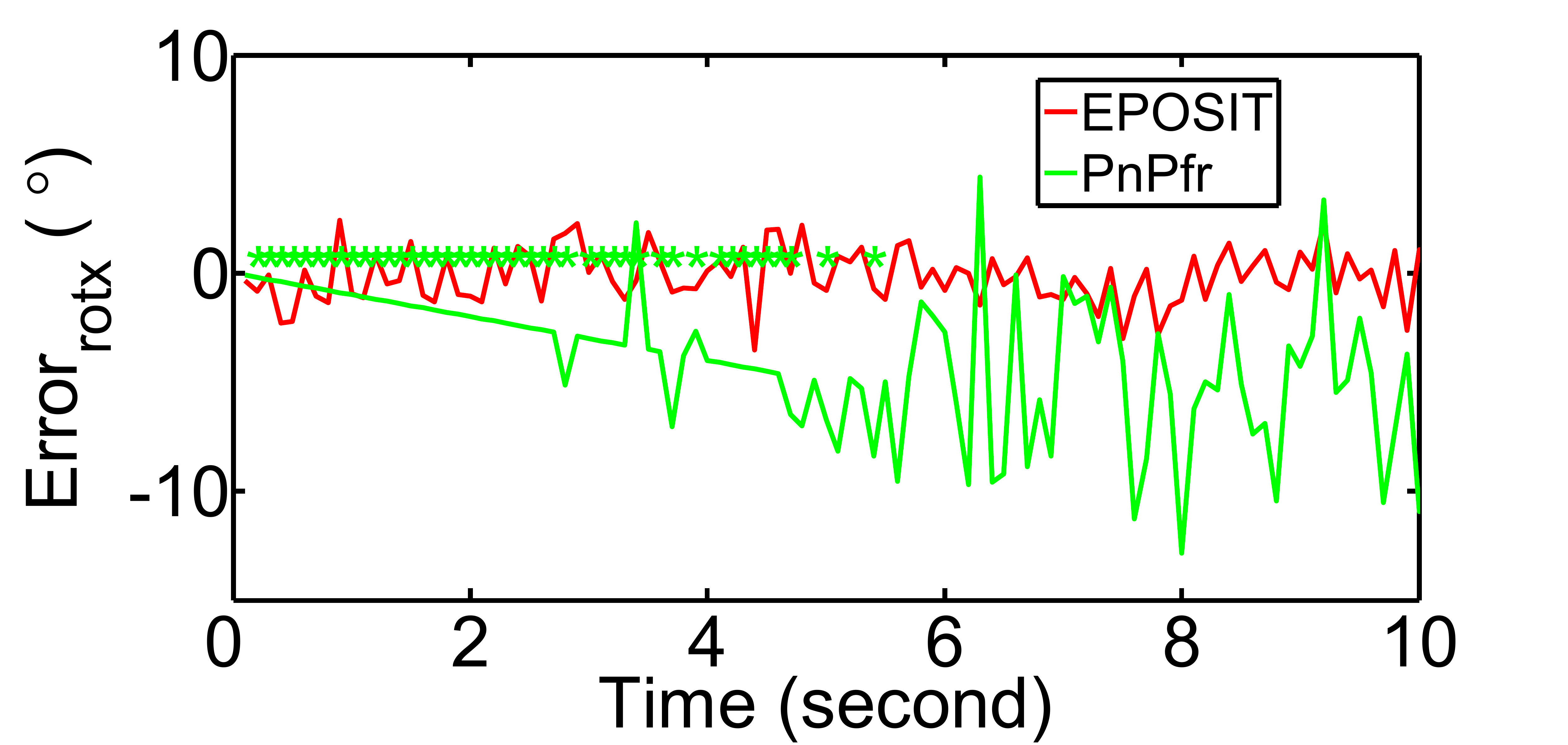} \label{fig8:a}
    }
\vspace{-0.45em}
    \subfigure[]
    {
        \includegraphics[width=1.58in]{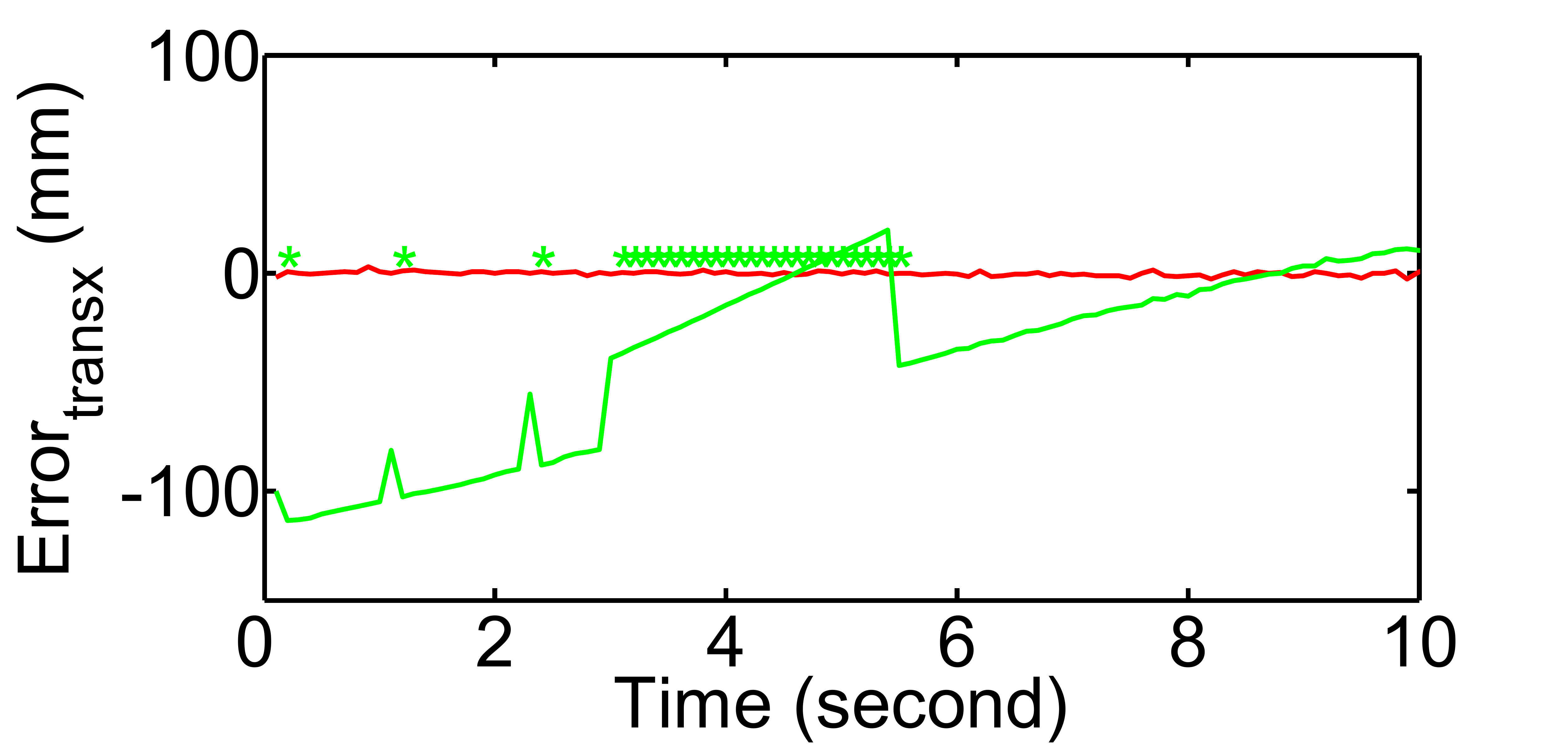}\label{fig8:b}
    }
\vspace{-0.45em}
    \\
    \subfigure[]
    {
        \includegraphics[width=1.58in]{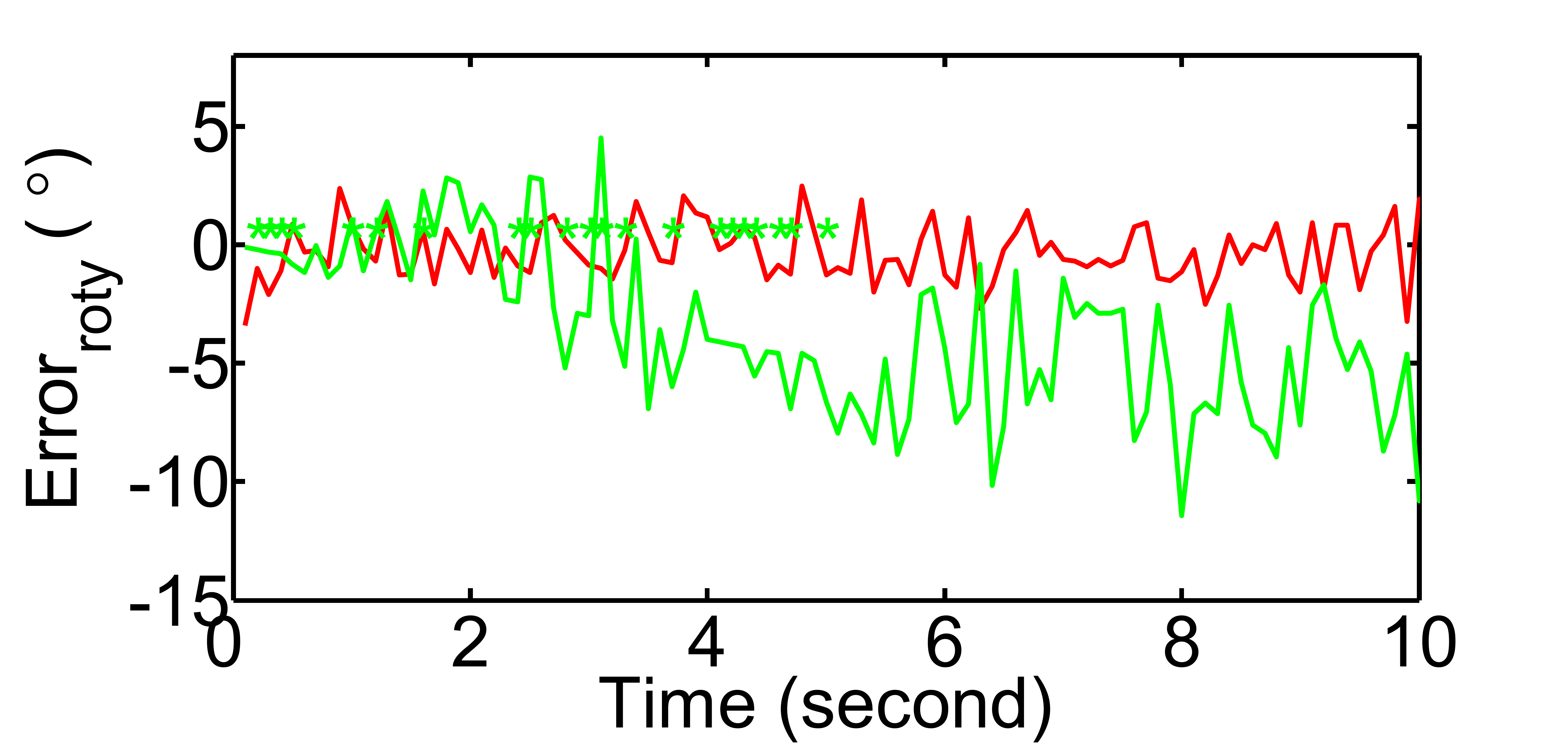} \label{fig8:c}
    }
\vspace{-0.45em}
    \subfigure[]
    {
        \includegraphics[width=1.57in]{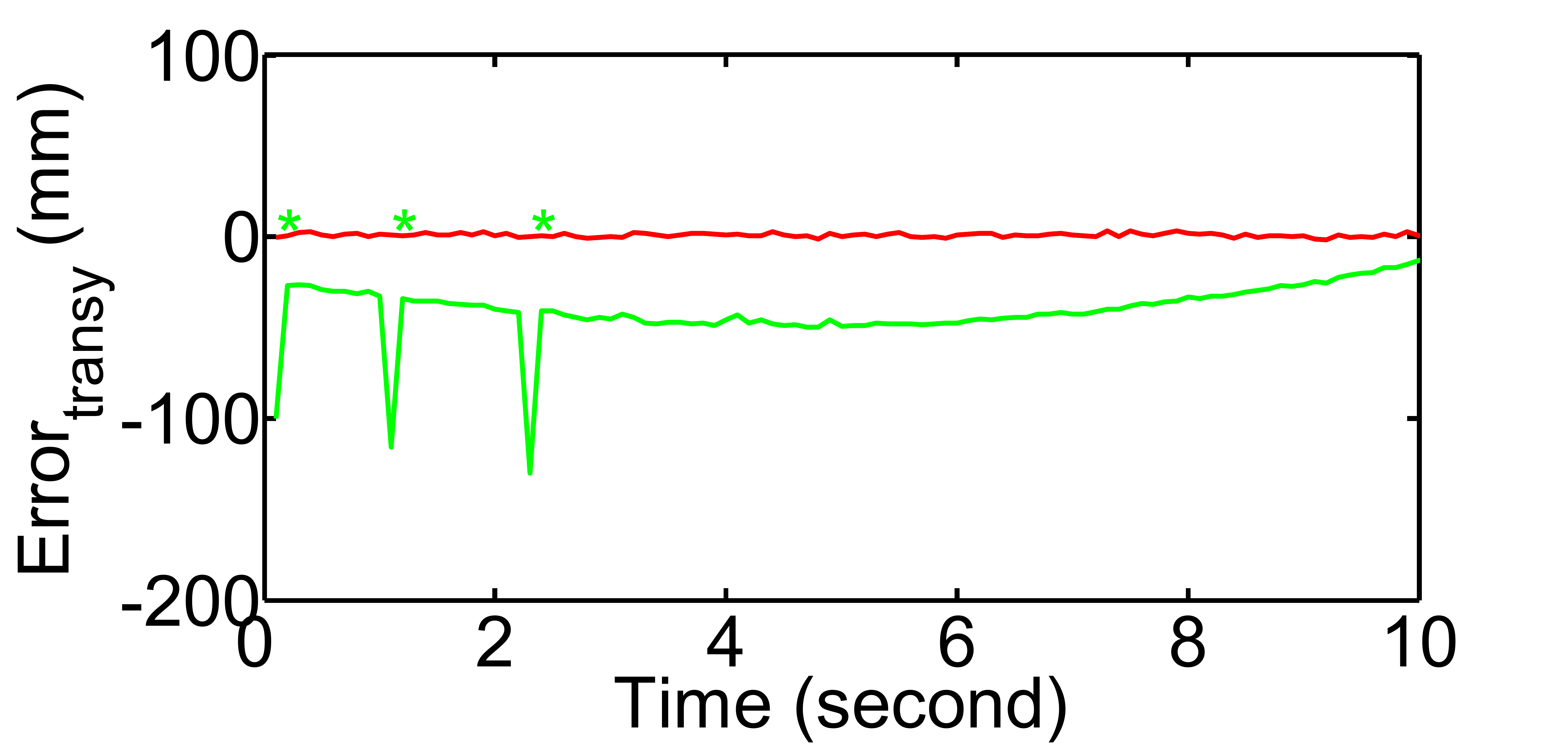}\label{fig8:d}
    }
\vspace{-0.45em}
 \\
    \subfigure[]
    {
        \includegraphics[width=1.58in]{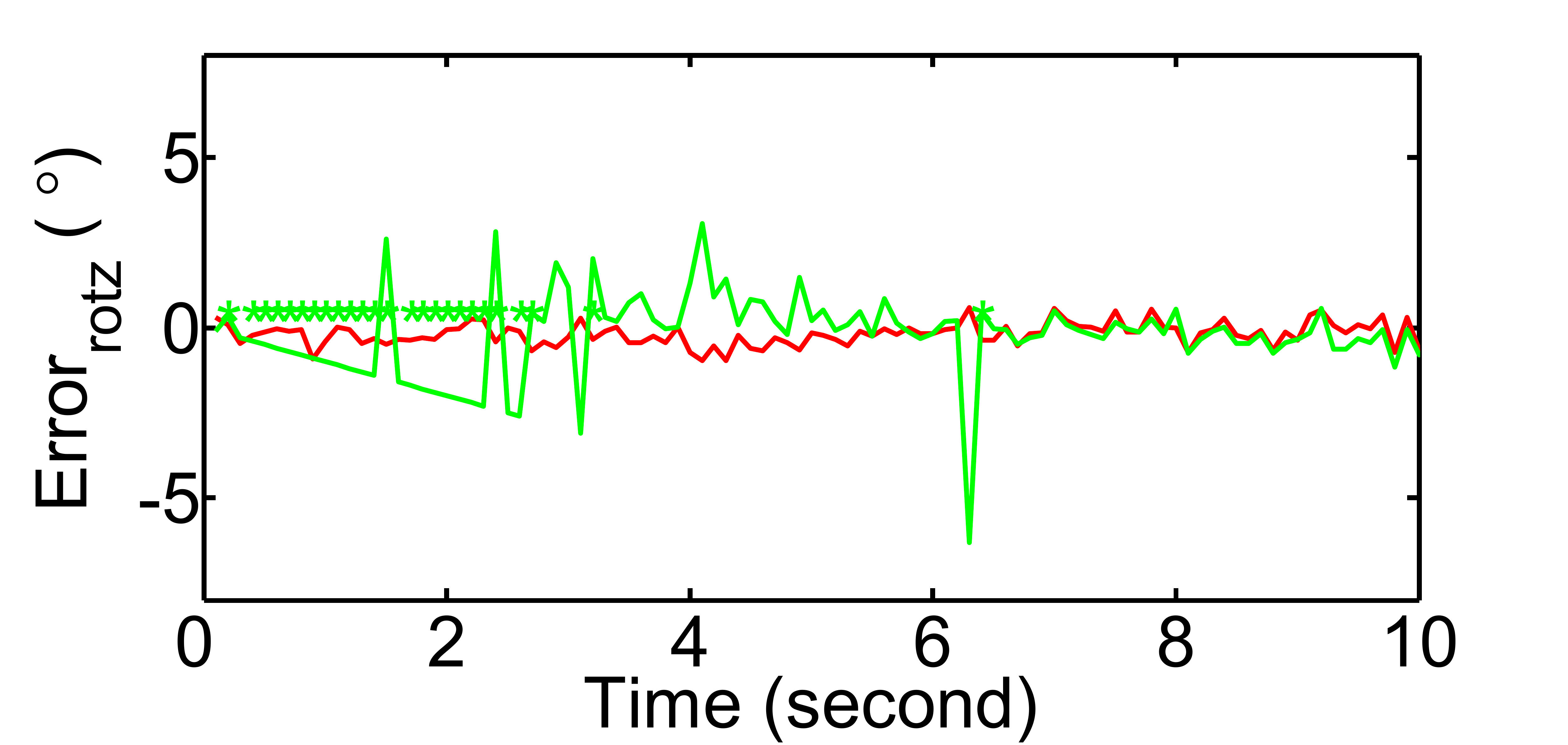} \label{fig8:e}
    }
    \subfigure[]
    {
        \includegraphics[width=1.58in]{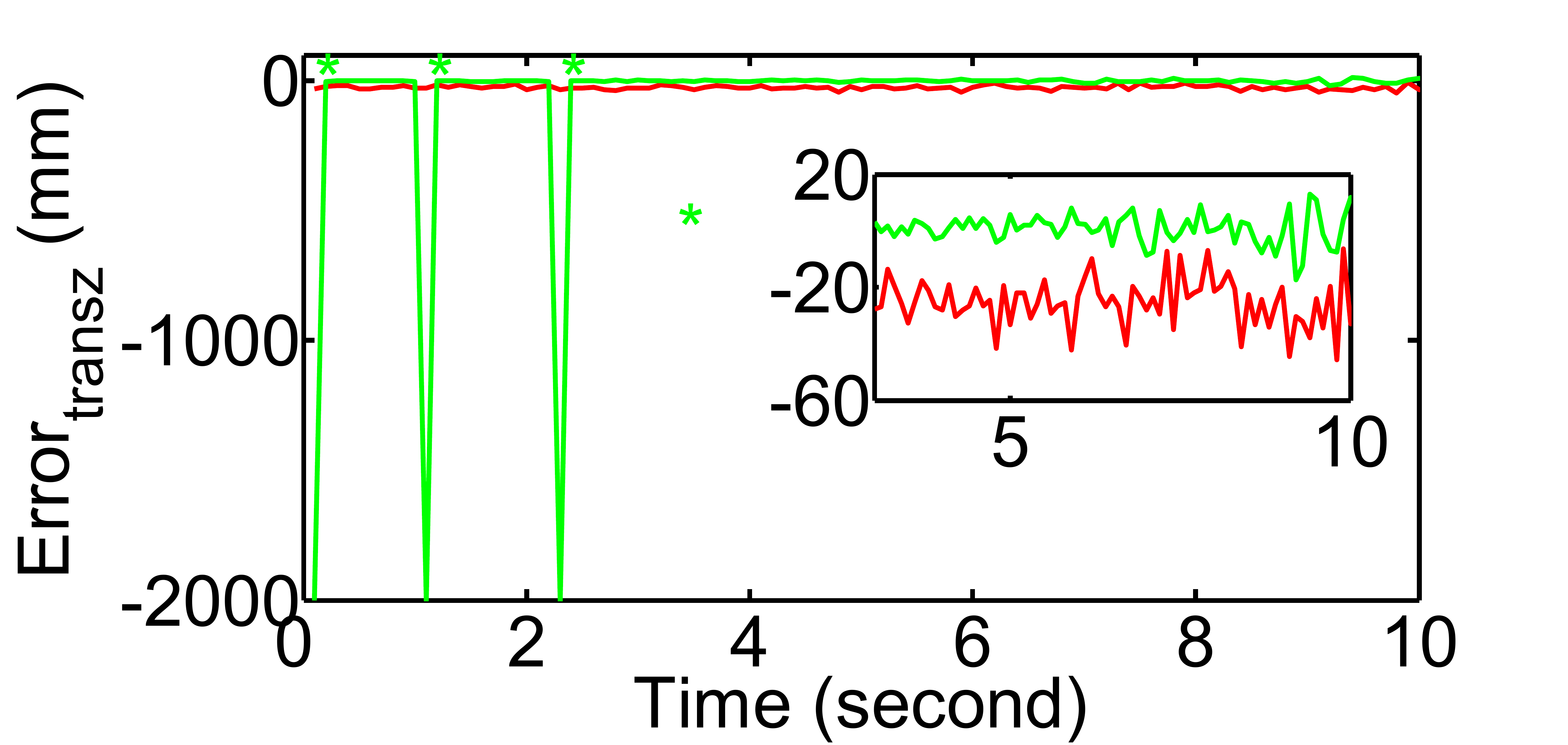}\label{fig8:f}
    }
   \caption{Comparison results for the fish-eye camera modelled by (\ref{eq5}). The left side represents estimation errors of the rotation, and the right side represents estimation errors of the translation.} 
    \label{20190902fig2}
\end{figure}

The performances of EPOSIT and PnPfr method for the fish-eye camera are also compared. In order to clearly clarify the comparison  results, only the fish-eye camera modelled by (\ref{eq5}) is employed to test these two methods, and their results are shown in Fig. \ref{20190902fig2}. From the left side, it can be seen that the error of EPOSIT is less than 3$^\circ$ for $x$ and $y$ axes while most errors of PnPfr are larger than 5$^\circ$. Although the error of $z$-axis is similar to ours, large vibrations exist in PnPfr. From the right side, the performance of EPOSIT is obviously better than PnPfr in $x$ and $y$ directions. However in $z$ direction, the error of EPOSIT is larger than the compared method (20$mm$ for EPOSIT and 10$mm$ for the compared method). The reason is $z$-axis value is aligned with the ground truth when recovering the scale of PnPfr result. 

From the above analysis, we can conclude that EPOSIT can precisely and stably estimate the pose of fish-eye cameras. 

{\setlength{\abovecaptionskip}{5pt}
\setlength{\belowcaptionskip}{-5cm}
\begin{figure}
    \centering
    \subfigure[]
    {
        \includegraphics[width=1.3in]{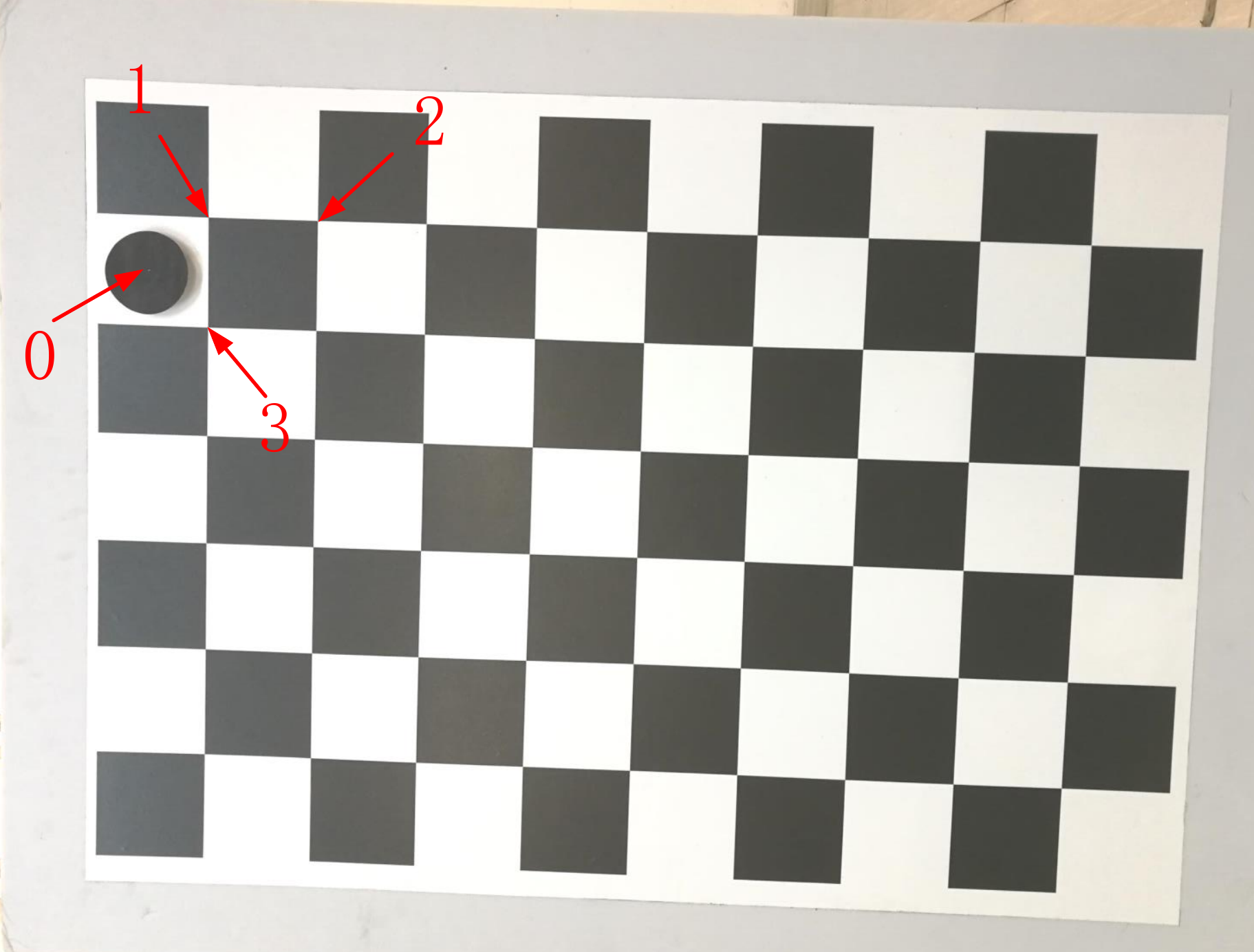}\label{fig7:a}
    }
    \subfigure[]
    {
        \includegraphics[width=1.4in]{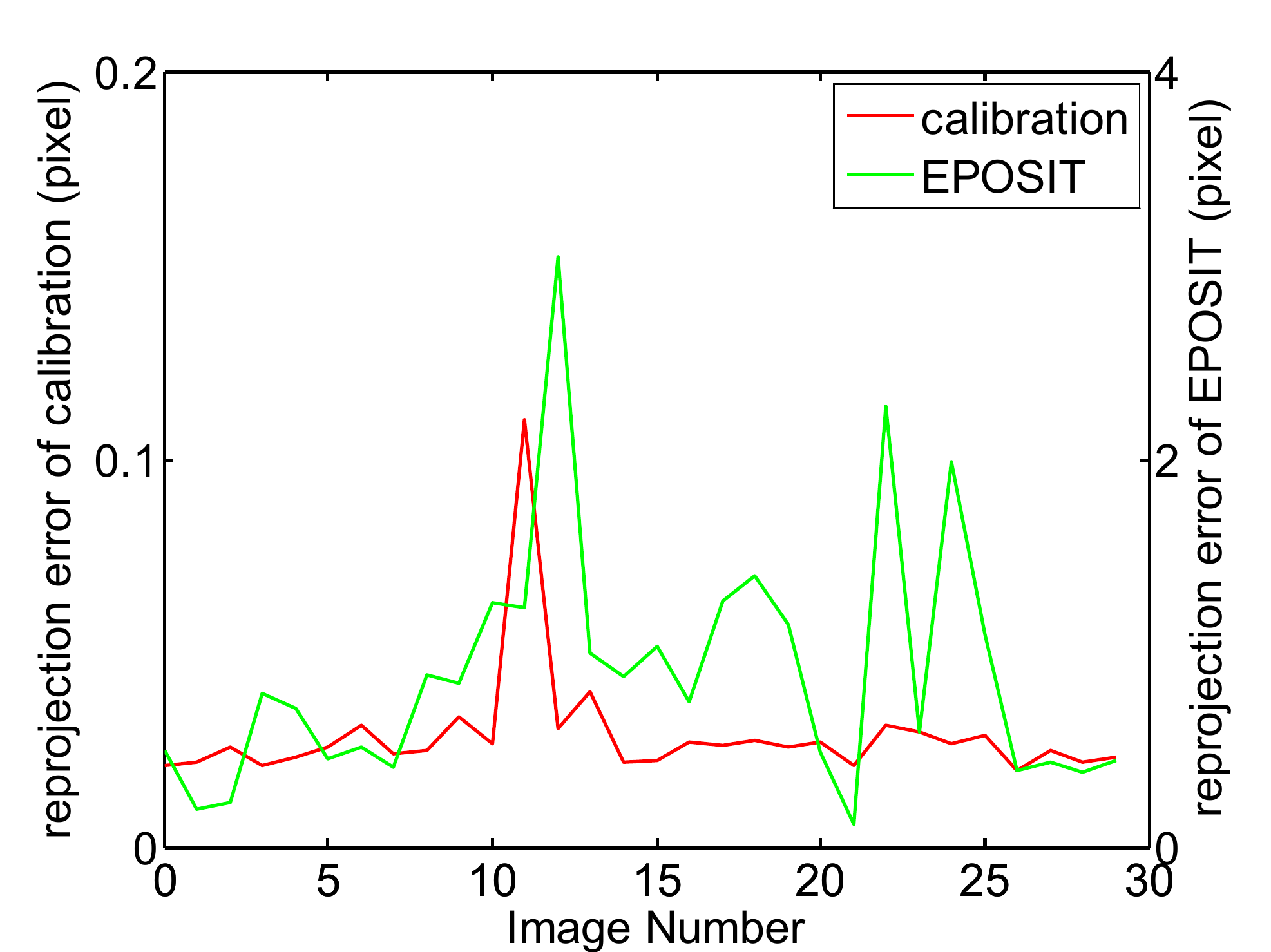} \label{fig7:b}
    }
    \caption{The chessboard and the image reprojection errors: (a) The chessboard used for the calibration and EPOSIT algorithm, and the height of the black circle to the chessboard is 10$mm$. (b) Reprojection errors of the calibration and EPOSIT.}
    \label{fig7}
\end{figure}}

{\setlength{\abovecaptionskip}{1pt}
\setlength{\belowcaptionskip}{-10cm}
\begin{figure}
    \centering
     \subfigure[]
    {
        \includegraphics[width=1.58in]{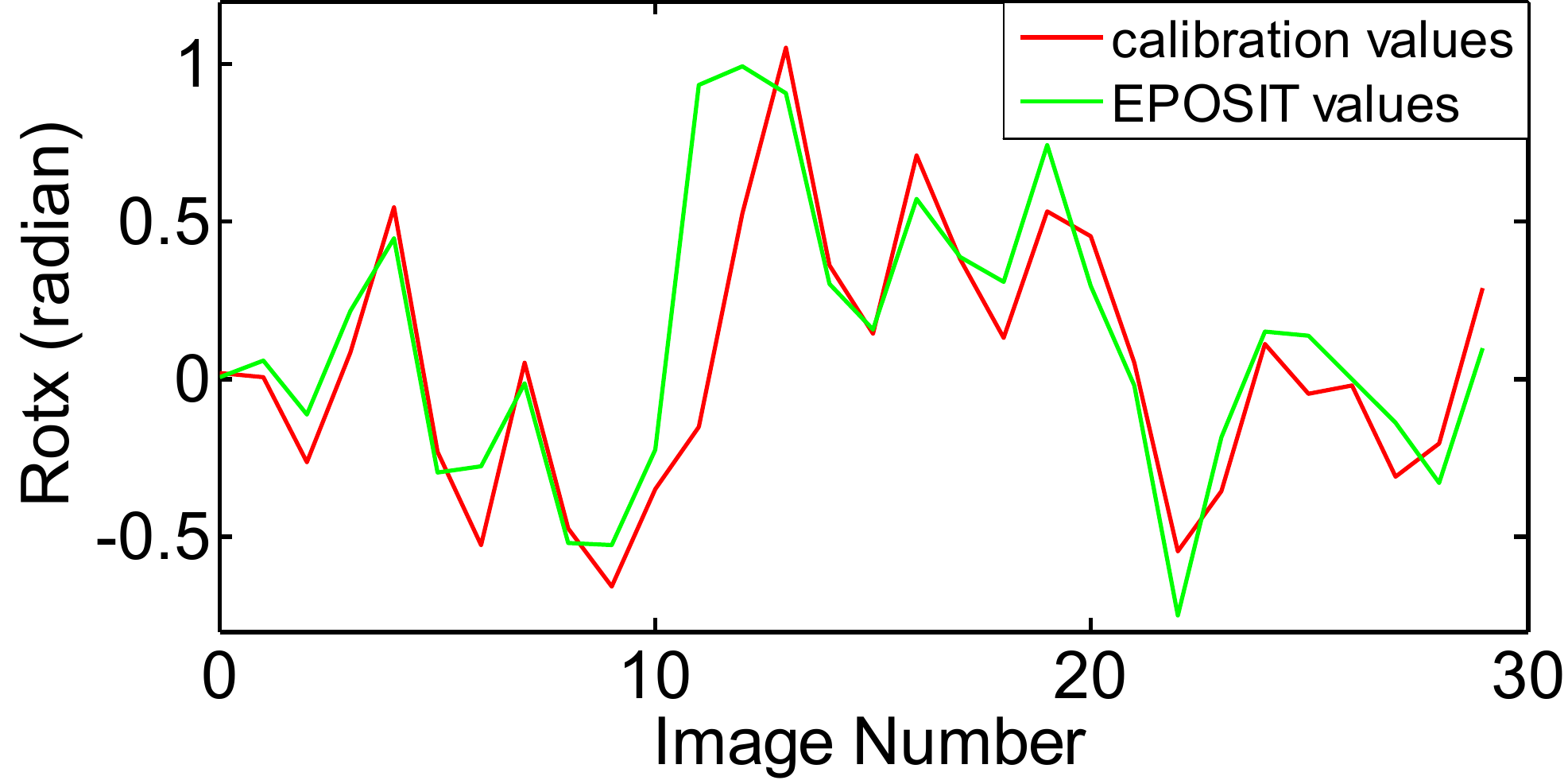} \label{fig8:a}
    }
\vspace{-0.45em}
    \subfigure[]
    {
        \includegraphics[width=1.58in]{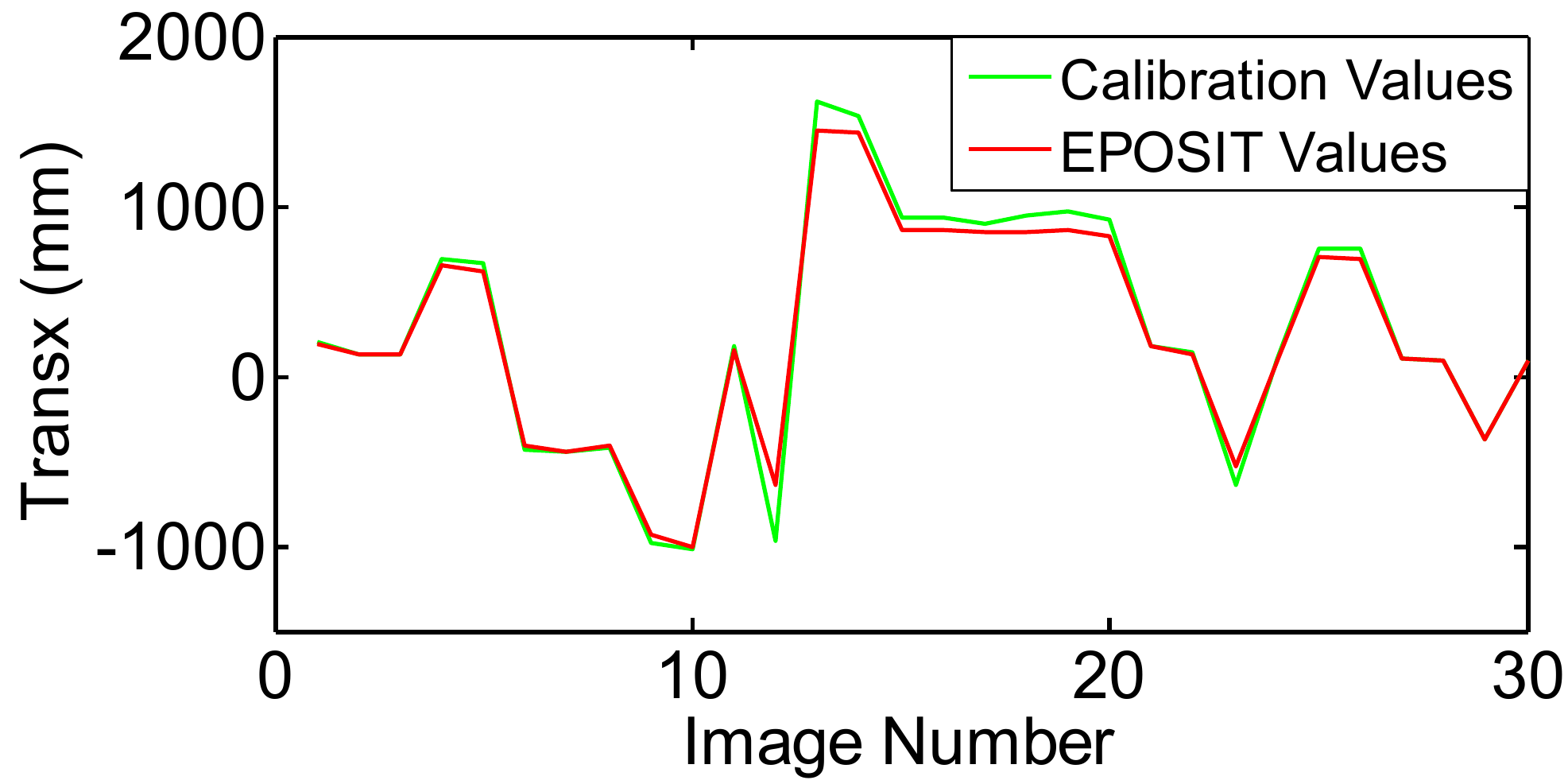}\label{fig8:b}
    }
\vspace{-0.45em}
    \\
    \subfigure[]
    {
        \includegraphics[width=1.58in]{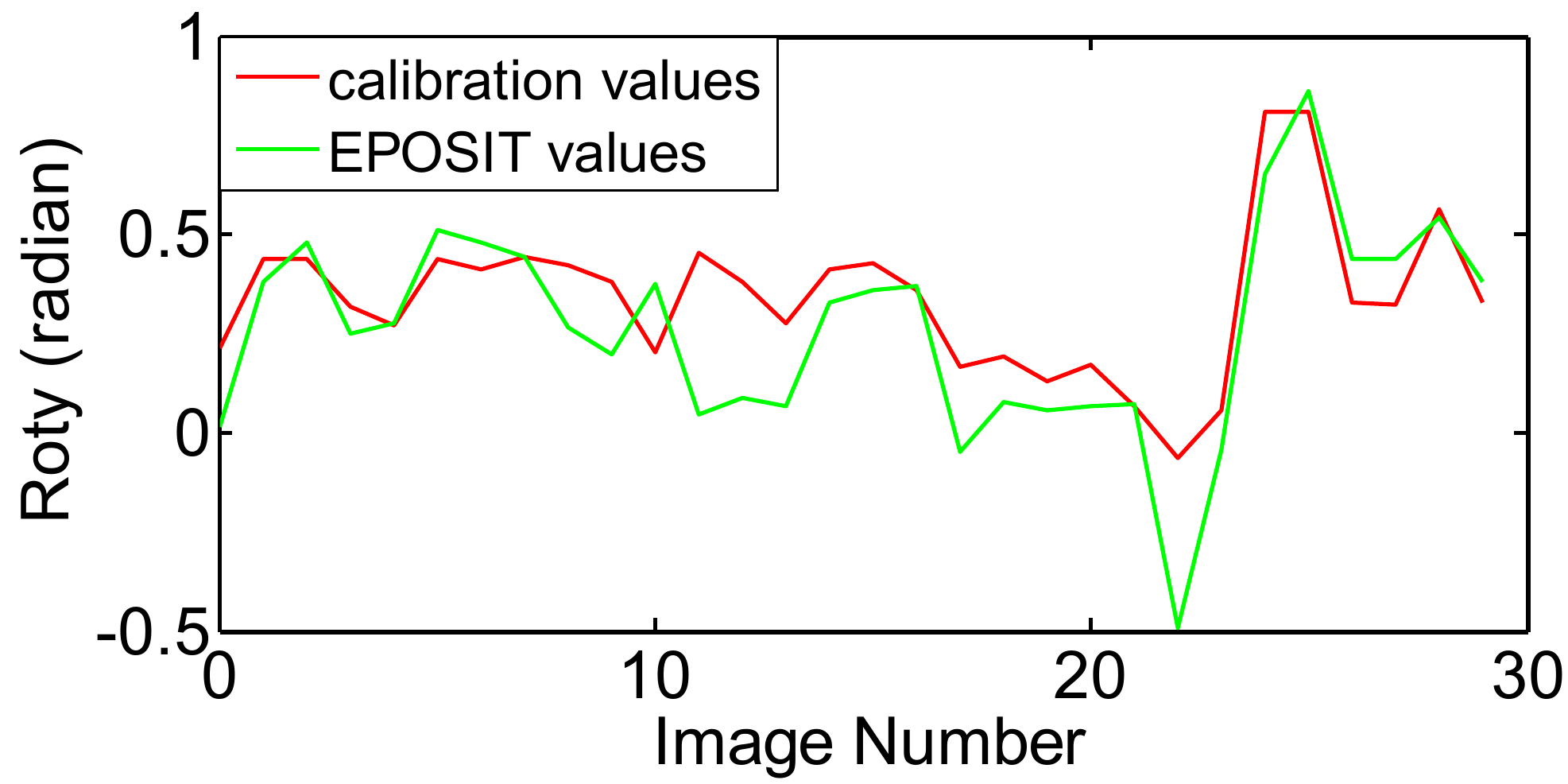} \label{fig8:c}
    }
\vspace{-0.41em}
    \subfigure[]
    {
        \includegraphics[width=1.57in]{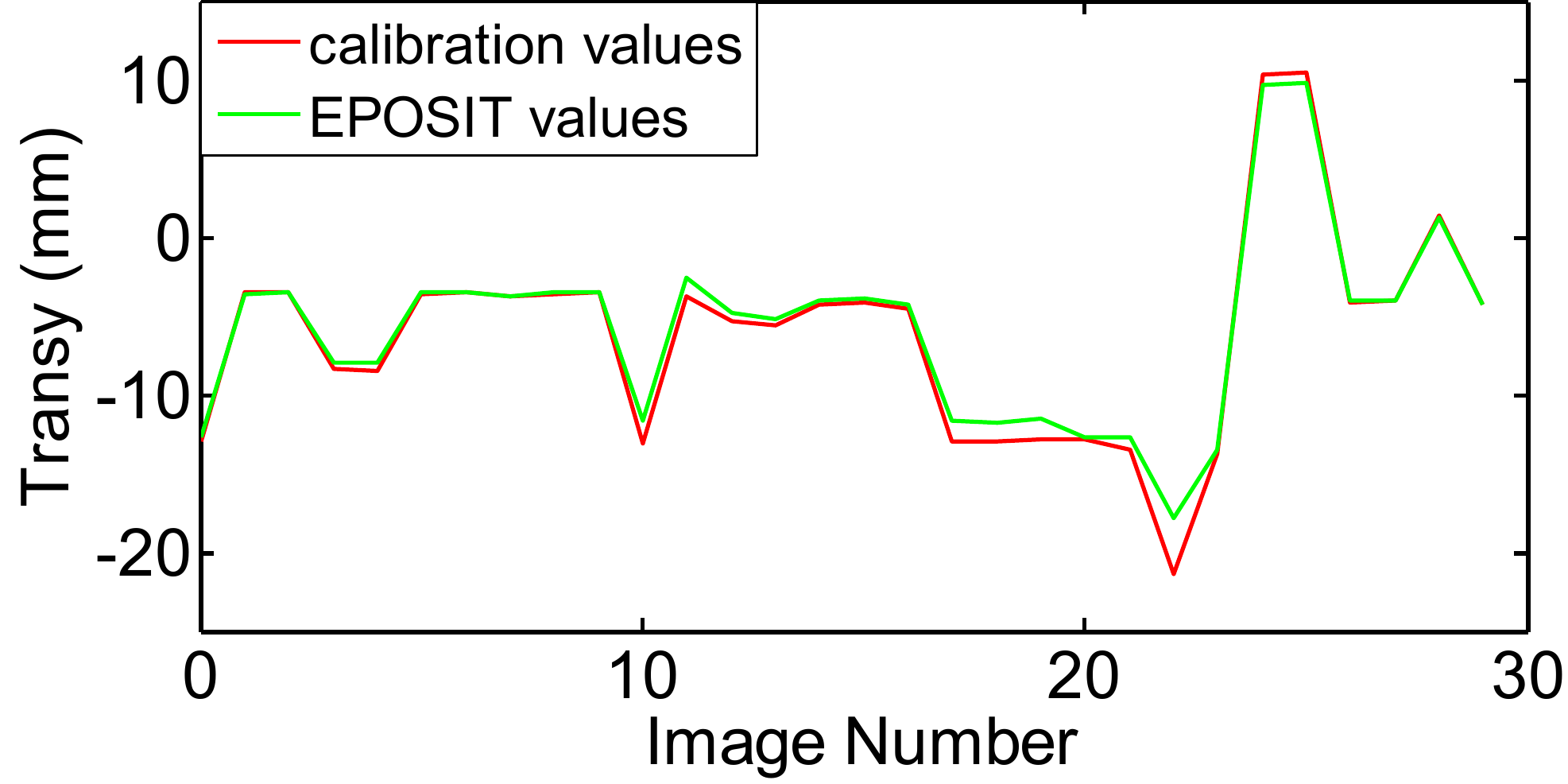}\label{fig8:d}
    }
\vspace{-0.41em}
 \\
    \subfigure[]
    {
        \includegraphics[width=1.58in]{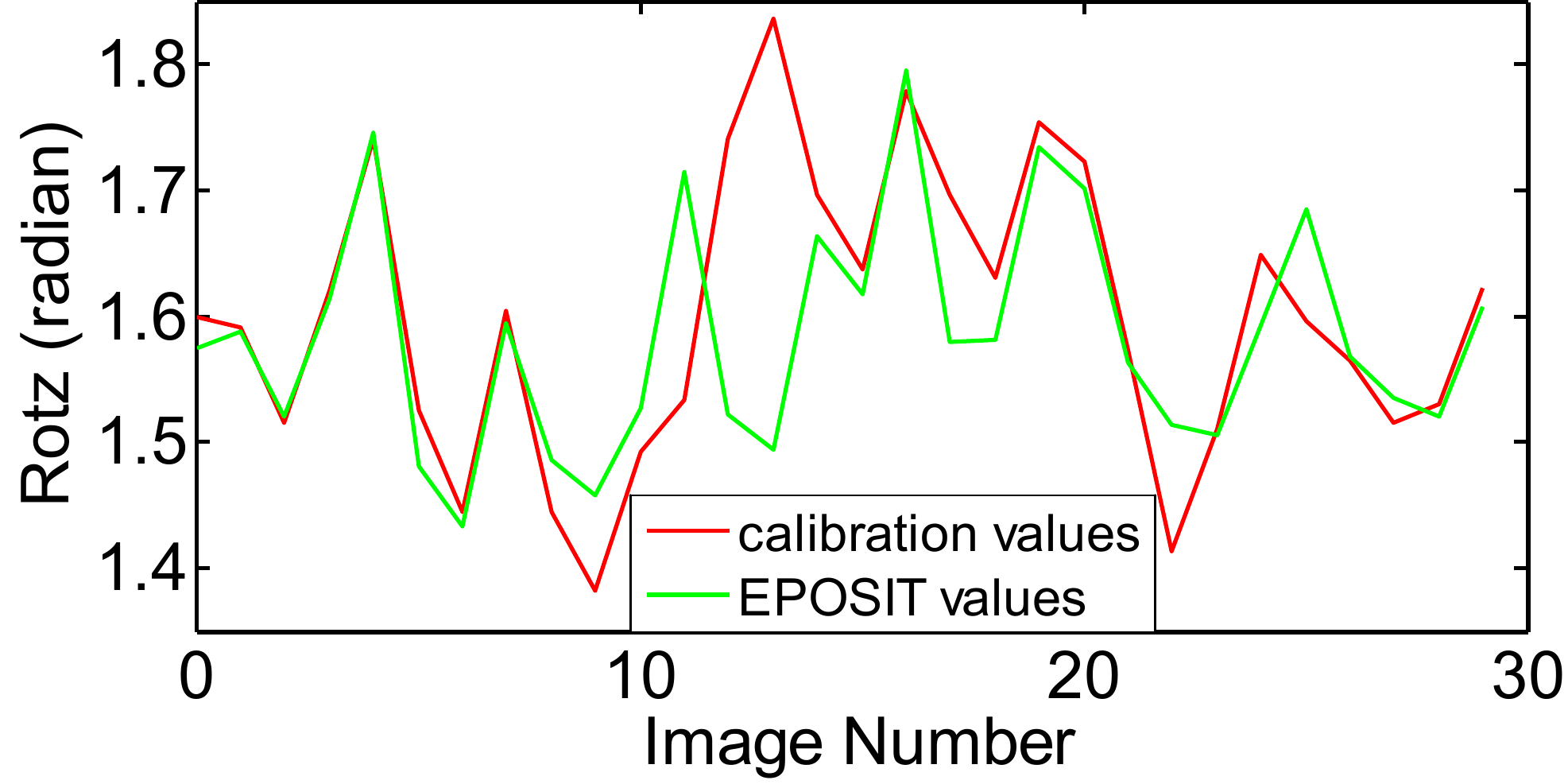} \label{fig8:e}
    }
    \subfigure[]
    {
        \includegraphics[width=1.58in]{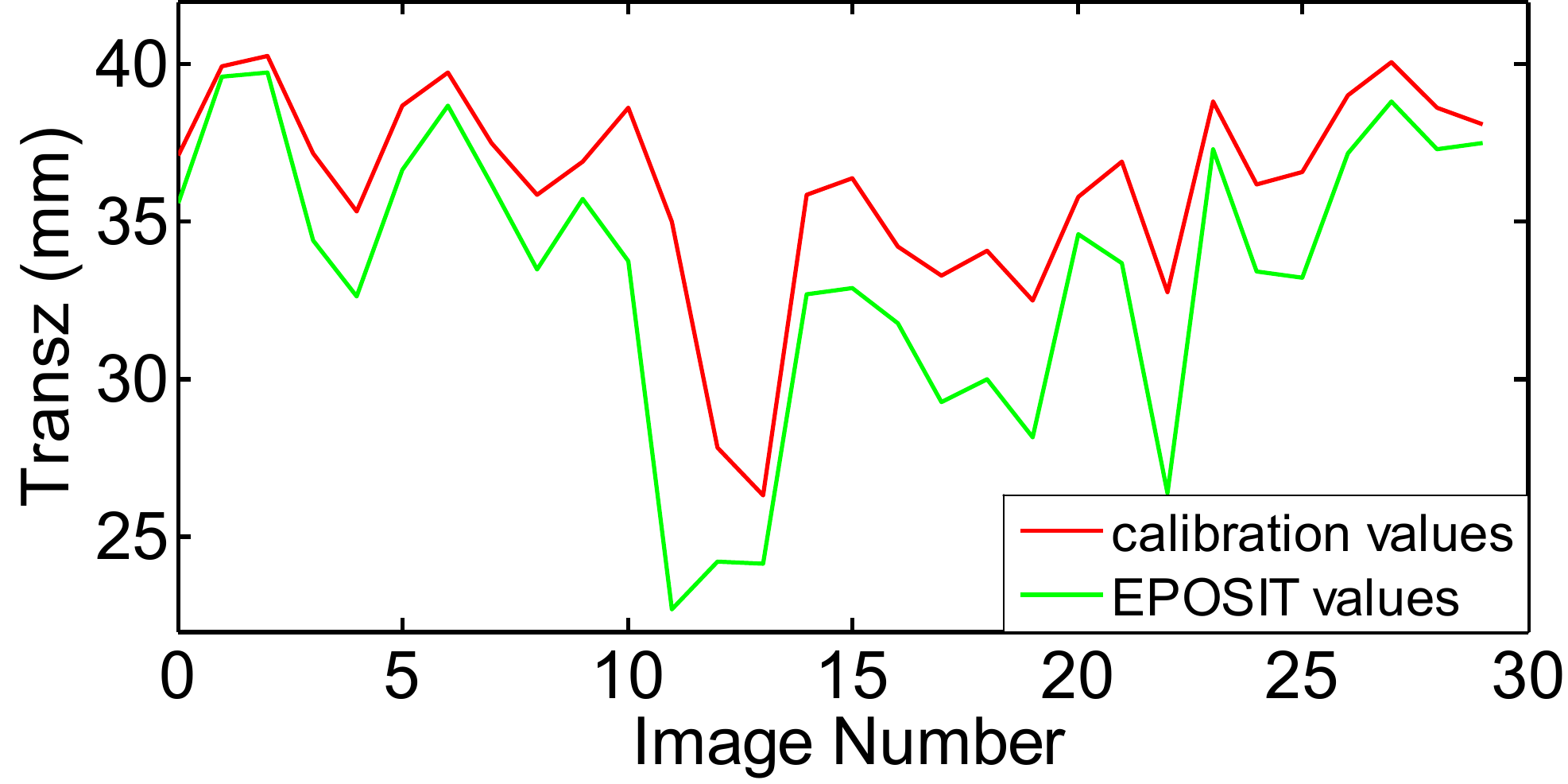}\label{fig8:f}
    }
   \caption{The comparison results: left side: the rotation estimation results around $x$, $y$ and $z$ axes respectively; right side: the translation estimation results in the $x$, $y$ and $z$ directions respectively.} 
    \label{fig8}
\end{figure}}
{\setlength{\abovecaptionskip}{1pt}
\setlength{\belowcaptionskip}{-1em}
\begin{figure} \centering 
 \subfigure[]
    {
        \includegraphics[width=1.58in]{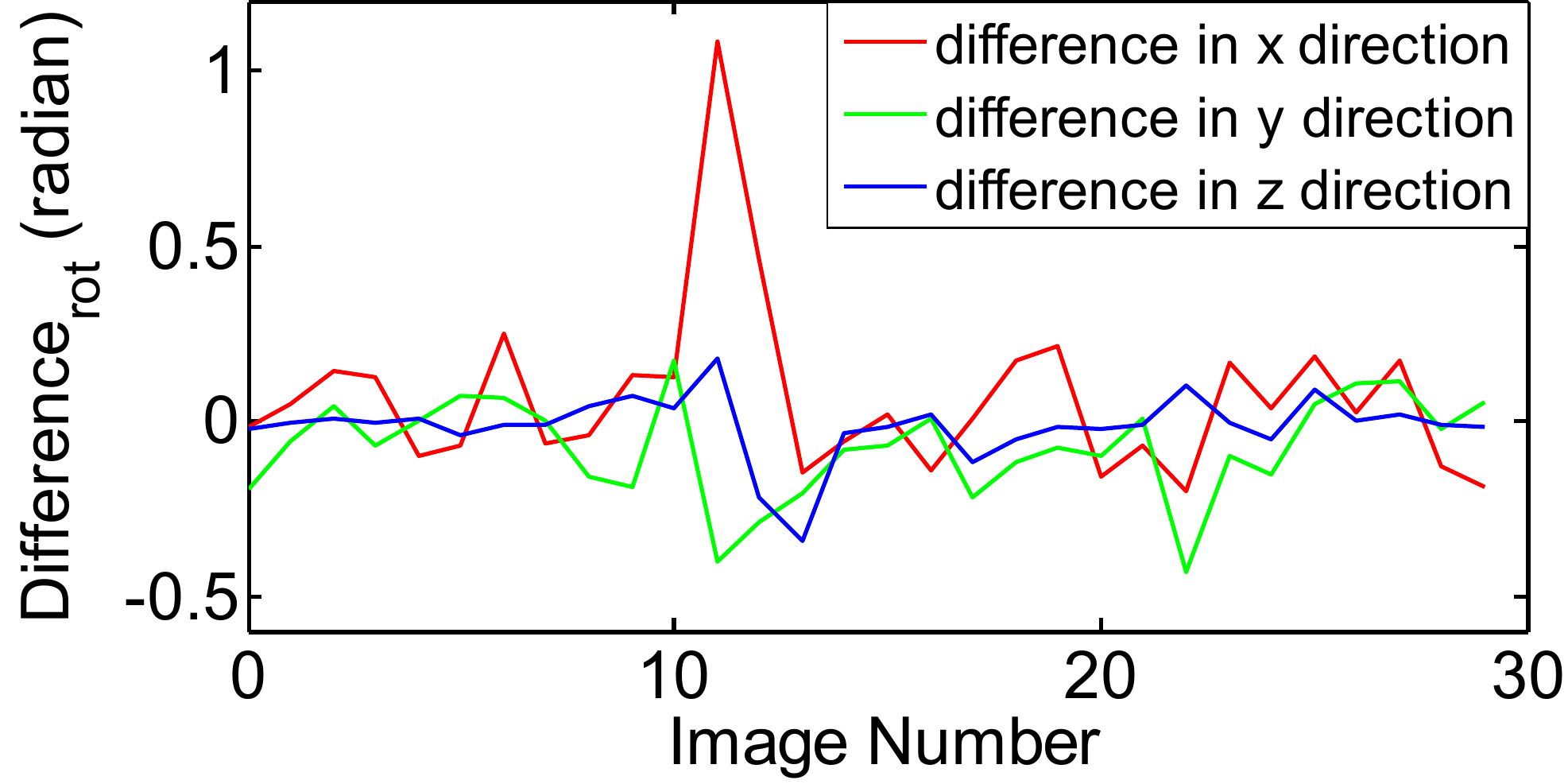} \label{fig9:a}
    }
    \subfigure[]
    {
        \includegraphics[width=1.58in]{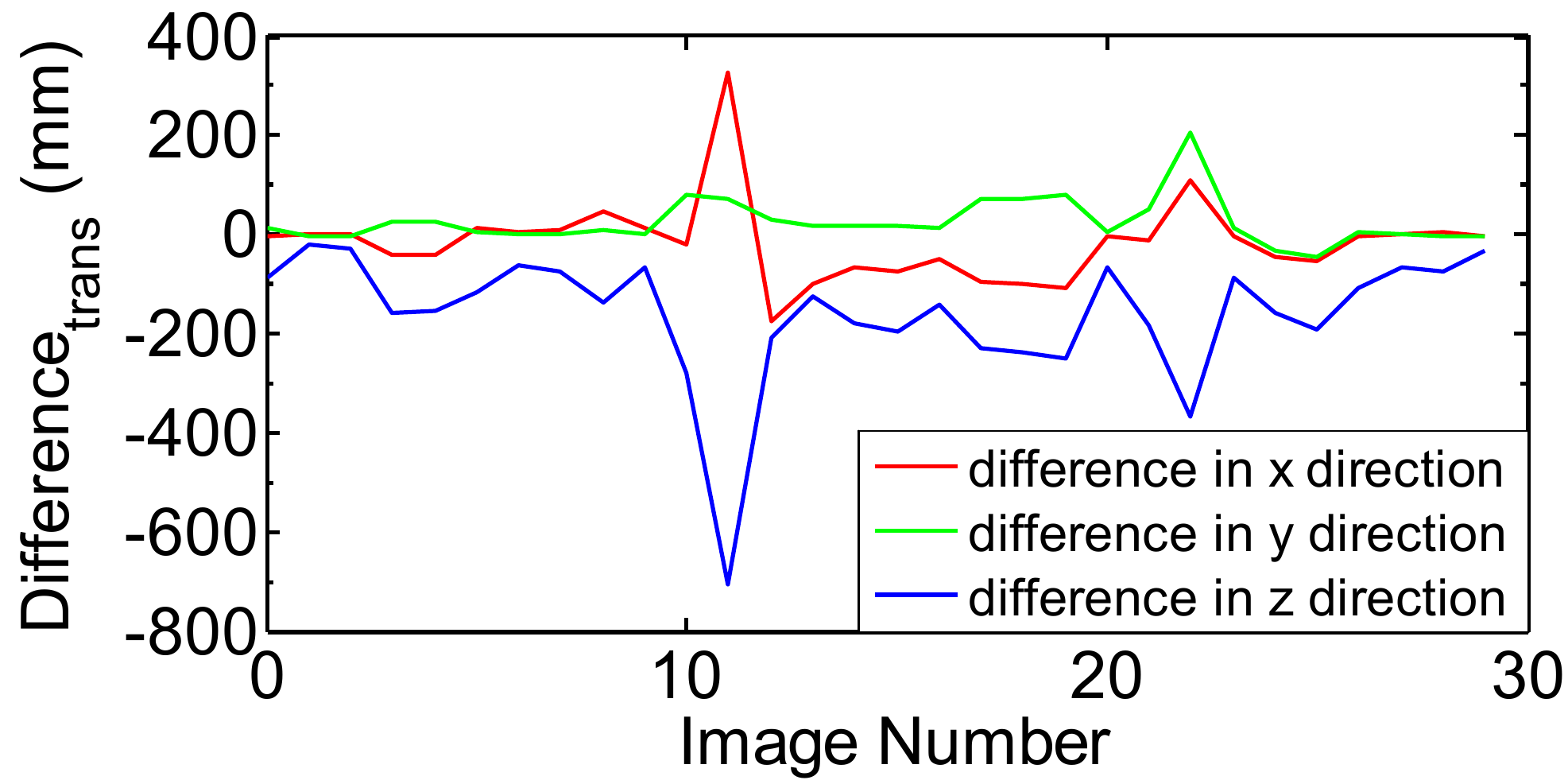}\label{fig9:b}
    }
\caption{The differences between EPOSIT estimation results and the calibration results. (a) the rotation results differences. (b) the translation results differences.} 
\label{fig9} 
\end{figure}}

\section{Experimental results}
The estimation results of EPOSIT algorithm are compared with calibration results in OpenCV3.0. In this experiment, a fish-eye camera modelled by (\ref{eq5}) is employed, whose extrinsic parameters are calibrated by a chessboard shown in Fig. \ref{fig7:a}. The focal length of the camera is 543 pixels, and the chessboard has 10$\times$7 squares whose size is 100$\times$100 $mm$. On the chessboard, a black circle is fixed in the middle of the first column of the second row, whose height to the chessboard is 10 $mm$. The object coordinate system $M_0uvw$ is coincide with the world coordinate system defined in OpenCV. The center of the black circle and the corners 1, 2 and 3 in Fig. \ref{fig7:a} are employed in EPOSIT algorithm to estimate the pose of the frame $M_0uvw$ relative to the camera coordinate system. The calibration results are acquired by 30 pictures. The estimation results of the calibration and EPOSIT algorithm, their differences and their reprojection error are respectively shown in Fig. \ref{fig8}, Fig. \ref{fig9} and Fig. \ref{fig7:b}. From Fig. \ref{fig8} and Fig. \ref{fig9}, we can see that the difference values are small except for the 11th, 13th and 22th images. From Fig. \ref{fig7:b}, it can be seen that, for these three images, their corresponding reprojection errors of the calibration results are larger than other pictures, which means that the calibrated extrinsic parameters of these three pictures far away from real values than other images. Therefore, the results of EPOSIT algorithm and calibration are different. For accurately calibrated images, the differences of the three rotation angles are all less than 0.2 radian according to Fig. \ref{fig9:a}. Most of the translation differences in the $x$ and $y$ directions are less than 50 $mm$ and less than 200 $mm$ in the $z$ direction according to Fig. \ref{fig9:b}. According to Fig. \ref{fig7:b}, the mean value of the reprojection errors of EPOSIT is 1.4 pixel without considering the 11th, 13th and 22th images. We run EPOSIT in an Intel Core i5-7400MQ @3.00 GHz computer with 8GB RAM, and 500 samples are tested. The average time-consumings of EPOSIT with and without feature detection are respectively 0.1$ms$ and 7.6$ms$.   

From the above analysis, the pose estimation results of EPOSIT are slightly different from calibration results. Therefore, EPOSIT is good enough to be used in most practical applications. Moreover, the results of EPOSIT algorithm can be obtained by one image in real-time, and better estimation performance can be achieved with more feature points.   

\section{CONCLUSION}
This paper presents a generic object pose estimation method which fits for both the pinhole camera and the fish-eye camera. The scaled orthographic projection model and the radially symmetric projection model are used to derive the generic estimation expression, and the iterative method is employed to calculate the pose estimation results. Simulation results show that EPOSIT algorithm can precisely calculate the pose estimation results and keep results stable to the error of the principal point. Experimental results show that EPOSIT algorithm is precise enough for practical applications. Using more than four points is helpful to make the algorithm more robust to the image noises and the point detection errors, which will be studied in the next work.


\begin{thebibliography}{00}
\bibitem{29} J. A. Hesch and S. I. Roumeliotis, ``A Direct Least-Squares (DLS) Method for PnP'', IEEE International Conference on Computer Vision, pp. 383-390, 2011.
\bibitem{30} R. Horaud, B. Conio, O. Leboulleux, and B. Lacolle, ``An analytic Solution for the Perspective-4-Point Problem'', IEEE Conference on Computer Vision and Pattern Recognition, pp. 500–507, 1989.
\bibitem{31} V. Lepetit, F. Moreno-Noguer, and P. Fua, ``EPnP: An Accurate O(n) Solution to the PnP Problem'', International Journal of Computer Vision, vol. 81, no. 2, pp. 155–166, 2008. 
\bibitem{32} G. Reid, J. Tang, and L. Zhi, ``A Complete Symbolic-Numeric Linear Method for Camera Pose Determination'', International Symposium on Symbolic and Algebraic Computation, pp. 215–223, 2003. 


\bibitem{4} M. A. Abidi and T. Chandra, ``A New Efficient and Direct Solution for Pose Estimation Using Quadrangular Targets: Algorithm and Evaluation'', IEEE Transactions on Pattern Analysis and Machine Intelligence, vol. 17, no. 5, pp. 534-538, 1995.
\bibitem{5} M. Faessler, E. Mueggler, K. Schwabe, \emph{et al}., ``A Monocular Pose Estimation System Based on Infrared LEDs'', International Conference on Robotics and Automation, pp. 907-913, 2014.
\bibitem{26} L. Kneip, D. Scaramuzza, R. Siegwart, ``A Novel Parameterization of the Perspective-Three-Point problem for a direct computation of absolute camera position and orientation'', IEEE International Conference on Computer Vision and Pattern Recognition, pp 2969–2976, 2011.
\bibitem{6} R. M. Haralick, H. Joo, C. Lee, \emph{et al}., ``Pose Estimation from Corresponding Point Data'', IEEE Transactions on Systems, Man, and Cybernetics, vol. 19, no. 6, pp. 1426-1446, 1989.
\bibitem{7} R. M. Haralick, C. Lee, K. Ottenberg, \emph{et al}., ``Review and Analysis of Solutions of the Three Point Perspective Pose Estimation Problem'', International Journal of Computer Vision, vol. 13, no. 3, pp. 331-356, 1994.
\bibitem{8} M. Roy, R. A. Boby, S. Chaudhary, \emph{et al}., ``Pose Estimation of Texture-less Cylindrical Objects in Bin Picking Using Sensor Fusion'', International Conference on Intelligent Robots and Systems, pp. 2279-2284, 2016.
\bibitem{9} E. Mair, K. H. Strobl, M. Suppa, \emph{et al}., ``Efficient Camera-Based Pose Estimation for Real-Time Applications'', IEEE International Conference on Intelligent Robots and Systems, pp. 2696-2703, 2009.
\bibitem{10} O. Tahri, H. Araujo, Y. Mezouar, \emph{et al}., ``Efficient Decoupled Pose Estimation from a Set of Points'', IEEE International Conference on Intelligent Robots and Systems, pp. 1608-1613, 2014.
\bibitem{11} Y. Kuang and K. Astr\"om, ``Pose Estimation with Unknown Focal Length Using Points, Directions and Lines'', IEEE International Conference on Computer Vision, pp. 529-536, 2013.
\bibitem{12} C. P. Lu, G. D. Hager, and E. Mjolsness, ``Fast and Globally Convergent Pose Estimation from Video Images'', IEEE Transactions on Pattern Analysis and Machine Intelligence, vol. 22, no. 6, pp. 610-622, 2000.
\bibitem{13} F. M. Mirzaei and S. I. Roumeliotis, ``Globally Optimal Pose Estimation from Line Correspondences'', IEEE International Conference on Robotics and Automation, pp. 5581-5588, 2011.
\bibitem{14} D. Campbell, L. Petersson, L. Kneip, \emph{et al}., ``Globally-Optimal Inlier Set Maximization for Camera Pose and Correspondence Estimation'', IEEE Transactions on Pattern Analysis and Machine Intelligence, DOI: 10.1109/TPAMI.2018.2848650, 2018. 
\bibitem{15} D. F. Dementhon and L. S. Davis, ``Model-Based Object Pose in 25 Lines of Code'', International Journal of Computer Vision, vol. 15, no. 1-2, pp. 123-141, 1995.
\bibitem{16} M. Bujnak, Z. Kukelova, and T. Pajdla, ``A General Solution to the P4P Problem for Camera with Unknown Focal Length'', IEEE Conference on Computer Vision and Patter Recognition, 2008.
\bibitem{17} Y. Zheng, S. Sugimoto, I. Sato, \emph{et al}., ``A General and Simple Method for Camera Pose and Focal Length Determination'', IEEE Conference on Computer Vision and Patter Recognition, pp. 430-437, 2014.
\bibitem{18} T. Sattler, C. Sweeney, and M. Pollefeys, ``On Sampling Focal Length Values to Solve the Absolute Pose Problem'', European Conference on Computer Vision, pp. 828-843, 2014.
\bibitem{19} K. Josephson and M. Byr$\ddot{o}$d, ``Pose Estimation with Radial Distortion and Unknown Focal Length'', pp. 2419-2426, IEEE Conference on Computer Vision and Patter Recognition, 2009. 
\bibitem{20} Z. Kukelova, M. Bujnak, and T. Pajdla, ``Real-time solution to the absolute pose problem with unknown radial distortion and focal length'', IEEE International Conference on Computer Vision, pp. 2816-2823, 2013.
\bibitem{21} V. Larsson, Z. Kukelova, and Y. Zheng, ``Making Minimal Solvers for Absolute Pose Estimation Compact and Robust'', IEEE International Conference on Computer Vision, pp. 2316-2324, 2017.
\bibitem{22} M. Bujnak, Z. Kukelova, and T. Pajdla, ``New efficient solution to the absolute pose
problem for camera with unknown focal length and radial distortion'', Asian Conference on Computer Vision, pp. 11-24, 2010.
\bibitem{23} Z. Kukelova, J. Heller, M, Bujnak, \emph{et al}., ``Radial Distortion Homography'', pp. 639-647, IEEE Conference on Computer Vision and Patter Recognition, 2015. 
\bibitem{add1} V. Larsson, Z. Kukelova and Y. Zheng, ``Camera Pose Estimation with Unknown Principal Point'', IEEE International Conference on Computer Vision and Pattern Recognition, pp 2982–2992, 2018.
\bibitem{add2} G. Nakano, ``A Versatile Approach for Solving PnP, PnPf, and PnPfr Problems'', European Conference on Computer Vision, pp. 338-352, 2016. 
\bibitem{24} C. Wu, ``P3.5P: Pose Estimation with Unknown Focal Length'', pp. 2440-2448, IEEE Conference on Computer Vision and Patter Recognition, 2015. 
\bibitem{25} J. Kannala and S. S. Brandt, ``A Generic Camera Model and Calibration Method for Conventional, Wide-Angle, and Fish-Eye Lenses'', IEEE Transactions on Pattern Analysis and Machine Intelligence, vol. 28, no. 8, pp. 1335-1340, 2006. 
\bibitem{28} Z. Y. Hu and F. C. Wu, ``A Note on the Number of Solutions of the Noncoplanar P4P Problem'', IEEE Transactions on Patter Analysis and Machine Intelligence, vol. 24, no. 4, pp. 550-555, 2002.
\bibitem{33} Y. Zheng, Y. Kuang, S. Sugimoto, \emph{et al}., ``Revisiting the PnP problem: A Fast, General and Optimal Solution'', IEEE International Conference on Computer Vision, pp. 2344-2351, 2013.
\end{thebibliography}
\end{document}